\documentclass{seg}
\usepackage{amsmath,amsfonts}
\usepackage{array}
\usepackage{subfig}
\usepackage{textcomp}
\usepackage{url}
\usepackage{verbatim}
\usepackage{graphicx}
\usepackage{booktabs}
\usepackage{multirow}

\hypersetup{hidelinks}
\usepackage{amsthm,amssymb,mathtools}

\usepackage{pifont}

\begin{document}

\title{CoGameReg: Cooperative-Game-Based Collaborative Semi-Supervised Regression for Seismic Reservoir Property Prediction}

\author{%
  Zhiyu Guo$^{1}$\and
  Jiwei Jia$^{1,2}$\thanks{Corresponding author: jiajiwei@jlu.edu.cn} \and
  Lei Li$^{3,4}$\thanks{Corresponding author: lilei09@cnpc.com.cn}
}
\date{}

\maketitle

\begin{center}
    \footnotesize
    $^{1}$School of Mathematics, Jilin University, Changchun, China\\
    $^{2}$AI for Science and Engineering Center, Shenzhen Loop Area Institute, Shenzhen, China\\
    $^{3}$BGP Inc., China National Petroleum Corporation, Zhuozhou, China\\
    $^{4}$National Engineering Research Center of Oil \& Gas Exploration Computer Software, Zhuozhou, China\\
\end{center}

\begin{abstract}
Accurate estimation of reservoir properties from seismic data remains challenging when well-log labels are sparse but three-dimensional seismic observations are abundant. We present CoGameReg, a cooperative-game-based semi-supervised regression framework that uses contribution-weighted predictions of heterogeneous regressors to construct shared pseudo-labels. This design makes each regressor's influence explicit in the learning targets generated for unlabeled seismic samples under sparse well control. The Tri-Game Shapley mechanism combines a rank–MSE performance game with a pairwise consistency game to estimate nonnegative regressor weights from labeled training data. These weights remain fixed during collaborative training and are reused for final prediction fusion. Candidate pseudo-labels are screened using a common local k-nearest-neighbor error-reduction criterion before retraining. We evaluate CoGameReg on 10 public regression datasets, a synthetic three-dimensional P-wave velocity prediction task, and a field water-saturation application. The benchmark advantage is most pronounced at the 2.5\% labeled-data ratio and decreases as supervision increases. CoGameReg obtains the best blind-well metrics in the synthetic experiment. In the field application, predictions clipped to 0–100\% yield a pooled coefficient of determination of 0.5942, a root mean square error of 5.7441 percentage points, and a Pearson correlation of 0.8265 across three blind wells. CoGameReg has the highest Pearson correlation among the four field methods. The results support the use of CoGameReg to integrate sparse well labels with dense seismic attributes, with performance varying across metrics and well locations.
\end{abstract}

\noindent\textbf{Keywords:} seismic reservoir property prediction;
semi-supervised learning; multi-attribute seismic data;
cooperative game theory; reservoir characterization; quantitative interpretation.

\section{Introduction}

Seismic reservoir property prediction is an important problem in quantitative
geophysics and plays a critical role in reservoir characterization and
hydrocarbon exploration. Traditional model-driven approaches infer reservoir
properties through seismic inversion and rock-physics-based relationships,
providing valuable physical interpretations of subsurface structures and
properties. However, their performance can be affected by assumptions in
forward modeling, inversion non-uniqueness, and accumulated uncertainties
during multiple processing stages \cite{ref_aki,ref_tarantola,ref_zoeppritz}.

With the increasing availability of three-dimensional seismic surveys,
large volumes of seismic data can be obtained, whereas reliable labels derived
from well logs remain extremely limited because of the high cost of drilling
and logging. This imbalance provides a practical motivation for learning strategies that
can exploit unlabeled seismic observations under limited supervision. In particular, accurately predicting reservoir properties
between sparsely distributed wells remains a major challenge because seismic
responses are affected by complex geological structures, heterogeneous
reservoir conditions, and nonlinear relationships between seismic attributes
and petrophysical properties.

Recent advances in artificial intelligence have enabled data-driven approaches
for seismic interpretation and reservoir characterization.
Deep learning methods have been successfully applied to seismic denoising,
fault detection, seismic facies classification, lithology prediction, and
reservoir parameter inversion
\cite{ref_denoise,ref_fault,ref_facies,ref_inversion}.
For reservoir property prediction, recent studies have explored direct learning
from multi-attribute seismic data rather than relying solely on traditional
inversion workflows. Because different seismic attributes capture
complementary information related to lithology, structure, and fluid response,
multi-attribute learning provides an effective way to integrate heterogeneous
geophysical information. However, existing deep learning approaches still face challenges associated
with limited well labels and generalization across different geological
settings \cite{ref_multiseismic,ref_gas_prediction_dl,ref_selfsupervised_gas}.

Semi-supervised learning is therefore a natural candidate for seismic
reservoir prediction because it can combine sparse well-log labels with
abundant unlabeled seismic samples. More generally, semi-supervised learning has become
an important paradigm for improving predictive performance when labeled data
are scarce but unlabeled data are abundant \cite{ref1,ref2}. In regression
settings, where continuous annotations are often expensive to obtain,
semi-supervised regression (SSR) is particularly relevant \cite{ref11}.
The broader semi-supervised learning literature includes self-training
\cite{ref5,ref6}, consistency regularization \cite{ref7,ref8},
graph-based learning \cite{ref_graph_ssl}, kernel-based classification
\cite{ref13}, and co-training \cite{ref3}.

A representative co-training framework for regression is COREG
\cite{ref35}, which employs two differently configured
$k$-nearest-neighbor ($k$-NN) regressors. COREG provides a
pseudo-label exchange mechanism based on local mean squared
error reduction and explicitly promotes diversity through
different distance metrics and/or neighborhood sizes.
The original study also considers fitting linear regression
or support vector regression models to the augmented training
sets produced by the $k$-NN-based labeling procedure.

Committee-based regression \cite{ref37} and S2RMS \cite{ref_S2RMS}
already employ multiple regressors for semi-supervised learning.
Within such settings, deciding which pseudo-labeled samples to accept
and deciding how to combine the regressors' predictions are related
but distinct design questions. This motivates investigating
contribution-aware weighting for shared pseudo-label construction
alongside local candidate screening.

A related challenge lies in the reliable exploitation of unlabeled data.
Prior studies have shown that unlabeled data may degrade performance when the
semi-supervised assumption is violated or when pseudo-labels are inaccurate
\cite{ref14,ref15,ref16,ref17}. In co-training regression, this issue is
particularly acute because pseudo-label quality depends not only on the
accuracy of individual regressors, but also on their interaction structure.
Representative SSR approaches address this risk through local error
reduction \cite{ref35}, self-paced selection with a safe prediction
mechanism \cite{ref39}, and confidence screening based on embedding
similarity and prediction dispersion \cite{ref_S2RMS}.
Alongside these sample-selection mechanisms, this work investigates
how estimated regressor contributions can determine the weights
used to construct shared pseudo-labels.

Motivated by these considerations, we propose CoGameReg (Cooperative-Game-Based Collaborative Semi-Supervised Regression) for seismic reservoir property prediction under sparse well control. CoGameReg treats heterogeneous regressors as players and estimates their contribution-aware weights once from the labeled training data. The Tri-Game mechanism combines coalition-level rank agreement and mean squared error with pairwise prediction consistency. The resulting weights determine the shared candidate pseudo-labels throughout collaborative training and are reused for final prediction fusion. Candidate admission is evaluated using a common local error-reduction criterion on each regressor's current augmented training set.
The central design choice is to use contribution weighting when constructing the additional learning targets, so that model contribution estimates affect both semi-supervised training and final prediction. This choice is relevant to seismic reservoir prediction because limited well samples constrain the relationship between reservoir properties and extensive seismic attribute volumes. Different regressors can produce different estimates from these same inputs, and their relative influence determines the pseudo-labels used for subsequent learning. CoGameReg makes this influence explicit through fixed model weights and complements it with local candidate screening. The contribution lies in this integration within the collaborative learning procedure; neither model agreement nor local screening alone establishes the geological reliability of a pseudo-label.

The main contributions of this paper are summarized as follows:

\begin{enumerate}
    \item We introduce CoGameReg, a heterogeneous semi-supervised regression framework in which fixed Shapley-derived weights determine each regressor's influence on both shared pseudo-label construction and final prediction fusion.

    \item We formulate a Tri-Game valuation mechanism combining a mixed rank–MSE performance game with a pairwise consistency game. The resulting model weights are used together with a common k-NN-based local error-reduction criterion for candidate pseudo-label admission.

    \item We evaluate the framework on 10 public regression datasets, a synthetic three-dimensional P-wave velocity prediction task, and a field water-saturation application. The seismic evaluations distinguish quantitative prediction at held-out wells from qualitative spatial consistency of the three-dimensional results.
\end{enumerate}

The experimental results show that CoGameReg is competitive with classical
co-training, self-training, ensemble, and recent semi-supervised baselines
under limited supervision. Its advantage on the public benchmarks is most
evident at the lowest labeled-data ratio. The synthetic and field applications
further demonstrate its applicability to seismic reservoir property prediction
with sparse well constraints, while also showing that prediction accuracy can
vary among blind-well locations.

\section{Related Work}

We review the literature most closely related to this study from three
perspectives: deep learning for seismic reservoir prediction, semi-supervised
regression and co-training, and game-theoretic valuation in machine learning.

\subsection{Deep Learning for Seismic Reservoir Prediction}

Deep learning has become an effective tool for seismic interpretation and
reservoir characterization. Convolutional neural networks and attention-based
architectures have been widely investigated for seismic denoising, fault
detection, seismic facies classification, lithology prediction, and reservoir
parameter inversion
\cite{ref_denoise,ref_fault,ref_facies,ref_inversion}. These studies
demonstrate the capability of data-driven models to extract complex nonlinear
relationships from seismic observations.

For quantitative reservoir characterization, recent studies have further
explored direct prediction of reservoir-related properties from multi-attribute
seismic data. Multi-attribute learning integrates complementary information
from seismic amplitude, frequency, phase, impedance, and other derived
attributes, providing richer representations of subsurface geological
conditions. For example, deep learning approaches have been investigated for
lithology prediction and fluid-related property estimation using multi-seismic
information and attribute fusion
\cite{ref_multiseismic,ref_gas_prediction_dl}.

Recent studies further illustrate the range of learning strategies used for seismic reservoir characterization. A convolutionally embedded transformer has been applied to reservoir elastic-property prediction from field seismic and well-log data, with comparison against physics-based amplitude-versus-offset inversion \cite{Ahmed2026Transformer}. A multiresolution approach combining Haar wavelet compression, convolutional and recurrent neural networks, and transfer learning has also been developed for seismic reservoir monitoring \cite{Ahmed2026Wavelet}. These studies address learned seismic representations and elastic-property estimation. CoGameReg examines how heterogeneous regressor predictions are combined to form the pseudo-labels used for learning from unlabeled seismic attributes under sparse well control.

Self-supervised pre-training on unlabeled seismic data followed
by few-shot finetuning has also been investigated for gas-bearing
prediction \cite{ref_selfsupervised_gas}.
Limited well supervision nevertheless remains a central constraint.
The present work investigates collaborative SSR in which unlabeled
seismic samples are incorporated through contribution-weighted
pseudo-label generation and local candidate admission.

\subsection{SSR and Co-Training}
SSR aims to improve predictive performance by jointly exploiting a
small labeled set and a large amount of unlabeled data \cite{ref11}.
Representative approaches include safe prediction \cite{ref17},
semi-supervised deep kernel learning \cite{jean2018ssdk}, and
uncertainty-aware deep regression \cite{dai2023ucvme}.
Within co-training regression, COREG \cite{ref35} uses local error
reduction to select pseudo-labeled samples.
Related collaborative regression methods include committee-based
regression \cite{ref37}, random-forest-based co-training regression
\cite{ref40}, self-paced safe co-training \cite{ref39}, and
embedding-space learning with pseudo-label smearing in S2RMS
\cite{ref_S2RMS}.

In the broader semi-supervised learning literature, co-training
ideas have been extended to tri-training \cite{ref31},
self-paced co-training for classification \cite{ref18},
view- and label-consistency-based co-training \cite{ref32},
and deep co-training \cite{ref33,ref34}.
Applications also include medical imaging, remote sensing,
time series, and learning to rank \cite{ref42,ref44,ref46,ref49}.

The present work focuses on coupling coalition-based contribution
allocation with shared pseudo-label construction and model-indexed
local admission. Fixed Shapley-derived weights determine each
regressor's influence on the shared pseudo-labels and are reused
for final prediction fusion, while a common local error-reduction
functional is evaluated on each regressor's current augmented
training set.

\subsection{Game-Theoretic Valuation in Machine Learning}

Game-theoretic methods provide principled tools for quantifying marginal
contributions in multi-agent systems. The Shapley value
\cite{shapley1953value} is a classical rule for contribution allocation and
has been widely adopted in machine learning, including feature attribution
\cite{ref_shap} and data valuation
\cite{ghorbani2019data}. Related coalition-based ideas have also appeared in
unsupervised learning. For instance, BiGC \cite{ref_BiGC} formulates clustering
through cooperative-game objectives and characterizes structural coherence
via coalition utilities induced by pairwise similarity.

These studies illustrate applications of cooperative-game concepts
to feature attribution, data valuation, and clustering.
CoGameReg evaluates coalitions of regressors through a mixed
rank--MSE performance game and a pairwise consistency game.
The resulting Shapley-derived contributions are normalized and
combined into weights estimated once from the labeled data.
These weights remain fixed throughout collaborative training,
where they determine the composition of shared candidate
pseudo-labels, and are reused for final prediction fusion.

\section{The Proposed Framework}
In this section, we present the CoGameReg framework, including an
overview of the framework, the Tri-Game Shapley valuation mechanism,
and the collaborative pseudo-labeling procedure. Here, “Tri-Game” denotes three conceptual valuation games—ordinal, accuracy, and consensus. By Shapley additivity, the ordinal and accuracy games are combined computationally into a mixed performance characteristic function, whereas consensus is represented by a separate characteristic function.

\subsection{Framework Overview}

CoGameReg constructs a pool of $N$ regressors
\begin{equation}
\mathcal{S}=\{\mathcal{M}_i\}_{i=1}^{N}, \qquad \mathcal{M}_i \in \mathcal{H}_i,
\end{equation}
together with a weighted ensemble predictor whose weight vector lies on the probability simplex
\begin{equation}
f_{\mathbf{w}}(\mathbf{x})=\sum_{i=1}^{N} w_i \mathcal{M}_i(\mathbf{x}),
\qquad \mathbf{w}\in\Delta^{N-1},
\end{equation}
where $\mathcal{H}_i$ is the hypothesis class of
the $i$-th regressor, $\Delta^{N-1}=\{\mathbf{w}\in\mathbb{R}^N_{\ge 0}:\sum_{i=1}^{N} w_i=1\}$.
After initialization on the labeled set $\mathcal{D}_L$, the Tri-Game
Shapley valuation mechanism is applied once to estimate an initial
contribution-aware weight vector $\mathbf{w}^{(0)}\in\Delta^{N-1}$. The obtained weights are fixed throughout
collaborative training and reused for final fusion, i.e.,
$\mathbf{w}^{*}=\mathbf{w}^{(0)}$.

At iteration $t$, collaborative training uses $\mathcal{U}'_{t-1}\subseteq\mathcal{D}_U$ as the current unlabeled candidate pool.
For each candidate $\mathbf{x}_u\in\mathcal{U}'_{t-1}$, a pseudo-label is generated by the fixed contribution-aware ensemble
\begin{equation}
\bar{y}_u=\sum_{i=1}^{N} w_i^{(0)} \mathcal{M}_i^{(t-1)}(\mathbf{x}_u).
\end{equation}
Accepted
pseudo-labeled samples are inserted into model-specific training sets, and the
regressors are retrained in a regressor-specific batchwise manner. Figure~\ref{fig:framework}
illustrates the overall workflow of CoGameReg, and the detailed pseudocode is provided in Appendix~A of the supplementary
material.

\begin{figure}[!htbp]
    \centering
    \includegraphics[width=0.70\linewidth]{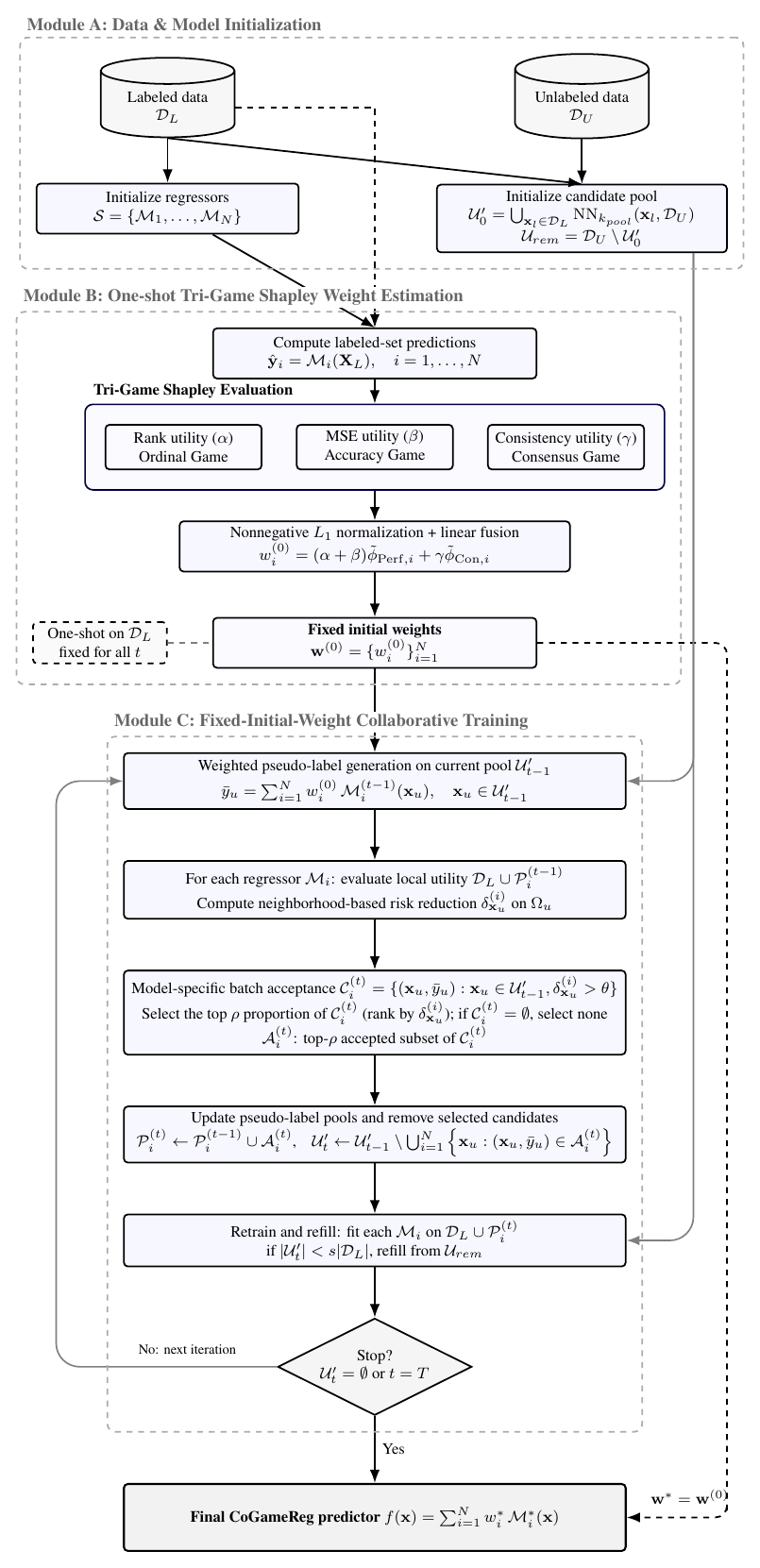}
    \caption{Overall workflow of the proposed CoGameReg framework.}
    \label{fig:framework}
\end{figure}

\subsection{Tri-Game Shapley Valuation Mechanism}
\label{sec:tri_game_shapley}

To estimate model-level authority, CoGameReg formulates the regressors in $\mathcal{S}$ as players in a transferable-utility cooperative
game. A characteristic function
\begin{equation}
v:2^{\mathcal{S}} \rightarrow \mathbb{R},
\end{equation}
assigns a utility to each coalition
$\mathcal{G}\subseteq\mathcal{S}$, with $v(\emptyset)=0$. For regressor
$\mathcal{M}_i\in\mathcal{S}$, its Shapley value is defined as
\begin{equation}
\phi_i(v)=
\sum_{\mathcal{G}\subseteq\mathcal{S}\setminus\{\mathcal{M}_i\}}
\frac{|\mathcal{G}|!(N-|\mathcal{G}|-1)!}{N!}
\left[v(\mathcal{G}\cup\{\mathcal{M}_i\})-v(\mathcal{G})\right].
\end{equation}
The Shapley value provides a principled basis for contribution allocation.

CoGameReg evaluates model utility from three complementary perspectives:
prediction accuracy, ordinal consistency, and cross-model consensus. These
perspectives are organized into two Shapley valuation components: a performance
component combining Rank and MSE utilities, and a consistency component
characterizing cross-model agreement. After computing the corresponding
Shapley-based contributions, we normalize and aggregate these components to
obtain the fixed initial ensemble weights.

\subsubsection{Consistency Assessment}

The consistency component evaluates cross-regressor consensus.
Inspired by coalition-based similarity formulations in cooperative-game
learning, we define the consistency utility of a coalition $\mathcal{G} \subseteq \mathcal{S}$ as
\begin{equation}
v_{\mathrm{Con}}(\mathcal{G})
=
\frac{1}{2}
\sum_{\mathcal{M}_i \in \mathcal{G}}
\sum_{\substack{\mathcal{M}_j \in \mathcal{G} \\ j \neq i}}
\exp\!\left(
-\frac{\|\hat{\mathbf y}_i-\hat{\mathbf y}_j\|_2^2}{2\sigma_{\mathrm{Con}}^2}
\right),
\label{eq:vcon}
\end{equation}
where $\hat{\mathbf y}_i$ denotes the prediction vector of regressor
$\mathcal{M}_i$ on the labeled set $\mathbf{X}_L$, and
$\sigma_{\mathrm{Con}}$ is set to the median value of the pairwise
Euclidean-distance matrix of these prediction vectors, plus
$\varepsilon=10^{-6}$.

Due to the pairwise-additive structure of $v_{\mathrm{Con}}$, the exact
Shapley value of regressor $\mathcal{M}_i$ is
\begin{equation}
\phi_{\mathrm{Con},i}
=
\frac{1}{2}
\sum_{\substack{\mathcal{M}_j\in\mathcal{S}\\j\ne i}}
\mathrm{Sim}(\mathcal{M}_i,\mathcal{M}_j),
\end{equation}
where $\mathrm{Sim}(\mathcal{M}_i,\mathcal{M}_j)$ denotes the Gaussian similarity
induced by their predictions on $\mathbf{X}_L$. Equivalently, $v_{\mathrm{Con}}(\mathcal{G})$ is the total edge
weight induced by coalition $\mathcal{G}$ in the
prediction-similarity graph. The corresponding Shapley allocation
assigns each regressor one-half of its incident edge weight. Because
this factor is common to all regressors, it cancels during
component-wise $L_1$ normalization; hence,
$\tilde{\phi}_{\mathrm{Con},i}$ is the normalized weighted-degree
centrality of $\mathcal{M}_i$ in this graph. This quantity
characterizes cross-model agreement rather than predictive
correctness.

\subsubsection{Performance Assessment}
The performance component evaluates predictive behavior with respect to the
labeled targets. Specifically, the Rank utility measures ordinal correctness,
whereas the MSE utility captures absolute prediction accuracy. These two
utilities capture complementary aspects of coalition-level predictive
behavior.

Instead of computing two separate Shapley values, we exploit the additivity
property of the Shapley value and define the following mixed characteristic
function for each nonempty coalition:
\begin{equation}
v_{\mathrm{Perf}}(\mathcal{G})
=
\alpha \cdot v_{\mathrm{Rank}}(\mathcal{G})
+
\beta \cdot v_{\mathrm{MSE}}(\mathcal{G}),
\qquad
\mathcal{G}\neq\emptyset,
\label{eq:vperf}
\end{equation}
where $\alpha,\beta\geq0$ and $\alpha+\beta\leq1$. Within the performance
game, the ratio between $\alpha$ and $\beta$ controls the balance between
ordinal and error-based criteria, whereas $\alpha+\beta$ determines the
overall contribution of the performance component in the final fusion.

Evaluating the Shapley value on this combined function yields the performance contribution $\phi_{\mathrm{Perf},i}$ through a single Shapley-value computation. The resulting performance contribution of regressor $\mathcal{M}_i$ is
\begin{equation}
\phi_{\mathrm{Perf},i}
=
\phi_i\!\left(v_{\mathrm{Perf}}\right).
\label{eq:phiperf}
\end{equation}

For a nonempty coalition
$\emptyset\neq\mathcal{G}\subseteq\mathcal{S}$, let
$\bar{\mathbf{y}}_{\mathcal{G}}$ denote its aggregated prediction on the labeled set, computed as the mean prediction of all regressors in the coalition:
\begin{equation}
\bar{\mathbf{y}}_{\mathcal{G}}
=
\frac{1}{|\mathcal{G}|}
\sum_{\mathcal{M}_i \in \mathcal{G}} \mathcal{M}_i(\mathbf{X}_L).
\end{equation}
The unweighted coalition average is used only for coalition utility evaluation,
thereby avoiding circular dependence between Shapley valuation and the weights to be
estimated. The two component utilities are defined as follows.
\begin{itemize}

\item Rank utility.
The ordinal agreement between the coalition prediction and the labeled
targets is measured using Spearman's rank correlation coefficient:
\begin{equation}
v_{\mathrm{Rank}}(\mathcal{G})
=
\frac{
1+\rho_s\!\left(
\mathbf{y}_{L},
\bar{\mathbf{y}}_{\mathcal{G}}
\right)
}{2}.
\end{equation}
Here, $\rho_s$ denotes Spearman's rank correlation coefficient. If the
standard deviation of either $\mathbf{y}_{L}$ or
$\bar{\mathbf{y}}_{\mathcal{G}}$ is smaller than $10^{-9}$, or if the
computed correlation coefficient is not finite, $\rho_s$ is set to zero.
Consequently, the corresponding rank utility is set to
$v_{\mathrm{Rank}}(\mathcal{G})=0.5$.

\item MSE utility.
The prediction error of a coalition is evaluated using an exponential
transformation of its mean squared error:
\begin{equation}
v_{\mathrm{MSE}}(\mathcal{G})
=
\exp\left[
-\frac{
\mathrm{MSE}\!\left(
\mathbf{y}_{L},
\bar{\mathbf{y}}_{\mathcal{G}}
\right)
}{
2b_{\mathrm{MSE}}^{\,2}
}
\right],
\end{equation}
where
\begin{equation}
b_{\mathrm{MSE}}
=
0.5\,\mathrm{Var}(\mathbf{y}_{L})
+\varepsilon,
\qquad
\varepsilon=10^{-6}.
\end{equation}
The value of $b_{\mathrm{MSE}}$ is computed once from the labeled targets
and is kept fixed for all coalition evaluations. The utility decreases
monotonically with coalition MSE and takes values in $(0,1]$.

\end{itemize}

\subsubsection{Contribution-Aware Model Fusion}

After computing the Shapley-based performance contribution and the consistency
contribution for each regressor, we combine them into a unified ensemble
weight. Since the performance Shapley values may be negative, we first apply a
zero-bounded $L_1$ normalization to preserve relative proportions while
filtering out detrimental marginal contributions. Specifically, the
performance Shapley values $\phi_{\mathrm{Perf},i}$ and consistency scores
$\phi_{\mathrm{Con},i}$ are separately normalized as
\begin{equation}
\begin{aligned}
\tilde{\phi}_{\mathrm{Perf},i}
&= \frac{\max(\phi_{\mathrm{Perf},i}, 0)}
{\sum_{k=1}^{N}\max(\phi_{\mathrm{Perf},k}, 0)}, \\
\tilde{\phi}_{\mathrm{Con},i}
&= \frac{\max(\phi_{\mathrm{Con},i}, 0)}
{\sum_{k=1}^{N}\max(\phi_{\mathrm{Con},k}, 0)}.
\end{aligned}
\label{eq:normalized}
\end{equation}
If the denominator of either component-wise normalization in
equation~\ref{eq:normalized} is zero, meaning that all zero-bounded component scores are
zero, the corresponding normalized component is set to the uniform distribution, namely, $\tilde{\phi}_{\mathrm{Perf},i}=1/N$ or
$\tilde{\phi}_{\mathrm{Con},i}=1/N$ for all $i$.

The fixed initial fusion weight is then obtained by aggregating these normalized
components:
\begin{equation}
w_i^{(0)}
=
(\alpha+\beta)\tilde{\phi}_{\mathrm{Perf},i}
+
\gamma \tilde{\phi}_{\mathrm{Con},i},
\end{equation}
where $\gamma=1-\alpha-\beta\geq0$, so that
$\alpha+\beta+\gamma=1$. Here,
$\alpha$ and $\beta$ determine the relative composition of the performance
game, while $\alpha+\beta$ specifies the total weight assigned to the
performance component relative to the consistency component in the final
fusion stage. Because both normalized contribution vectors lie on the probability
simplex and $(\alpha+\beta)+\gamma=1$, the fixed weight vector also
satisfies $\mathbf{w}^{(0)}\in\Delta^{N-1}$. Consequently, the fused
prediction lies within the pointwise range of the base-regressor
predictions; this convex-hull property does not impose a physical
bound unless the constituent predictions are themselves bounded.

Exact evaluation of the performance Shapley values requires enumeration of
$2^N$ coalitions. In the reported experiments, the ensemble contains five
regressors for the field application and six regressors for the benchmark and
synthetic experiments, and the fixed contribution weights are evaluated only
once before collaborative training. The exact performance valuation therefore requires 32 coalition evaluations for the field application and 64 for the benchmark and synthetic experiments. For larger ensembles, exact enumeration
can be replaced by Monte Carlo permutation sampling.

\subsection{CoGameReg Collaborative Training Procedure}
The proposed CoGameReg framework performs iterative collaborative expansion over a locality-aware candidate pool. In each iteration, pseudo-labeled candidates are evaluated and selected in a regressor-specific batchwise manner, incorporated into corresponding model-specific training sets, and then used to retrain the regressors; the candidate pool is refilled when needed.

\subsubsection{Hybrid Unlabeled Pool Initialization and Progressive Expansion}

To improve the reliability of early pseudo-label selection, CoGameReg adopts a
hybrid unlabeled-pool strategy that combines locality-aware initialization
with progressive exploration. At the start of training, we construct the
initial candidate pool $\mathcal{U}'_0$ by selecting, for each labeled sample
$\mathbf{x}_l\in\mathcal{D}_L$, its $k_{\mathrm{pool}}$ nearest neighbors from
$\mathcal{D}_U$:
\begin{equation}
\mathcal{U}'_0 = \bigcup_{\mathbf{x}_l \in \mathcal{D}_L}
\mathrm{NN}_{k_{\mathrm{pool}}}(\mathbf{x}_l, \mathcal{D}_U),
\end{equation}
where $\mathrm{NN}_{k_{\mathrm{pool}}}(\cdot)$ denotes the
$k_{\mathrm{pool}}$-nearest neighbor operator. The remaining unlabeled set is
denoted by
\begin{equation}
\mathcal{U}_{\mathrm{rem}}=\mathcal{D}_U\setminus\mathcal{U}'_0.
\end{equation}
As training proceeds, if the size of the updated candidate pool
$\mathcal{U}'_t$ falls below the target scale $s|\mathcal{D}_L|$, new samples
are added from $\mathcal{U}_{\mathrm{rem}}$ into $\mathcal{U}'_t$ and removed
from $\mathcal{U}_{\mathrm{rem}}$.

\subsubsection{Weighted Pseudo-Labeling, Model-Specific Acceptance, and Dynamic Batching}

Given the initialized regressor pool
$\mathcal{S}=\{\mathcal{M}_1,\dots,\mathcal{M}_N\}$, CoGameReg first estimates the
fixed initial ensemble weights $\{w_i^{(0)}\}_{i=1}^{N}$ using the Tri-Game
Shapley mechanism described in Section~\ref{sec:tri_game_shapley}. For each
candidate sample $\mathbf{x}_u \in \mathcal{U}'_{t-1}$, the pseudo-label is then
generated by the fixed weighted ensemble:
\begin{equation}
\bar{y}_u = \sum_{i=1}^{N} w_i^{(0)} \mathcal{M}_i^{(t-1)}(\mathbf{x}_u).
\label{eq:weighted_pseudo}
\end{equation}

With $\mathbf{w}^{(0)}$ fixed, perturbing only the prediction of
regressor $\mathcal{M}_i$ by $\varepsilon$ changes the shared
pseudo-label by
$\Delta\bar{y}_u
=
w_i^{(0)}\varepsilon.$
Thus, $w_i^{(0)}$ is the direct linear influence coefficient of
$\mathcal{M}_i$ in both candidate pseudo-label construction and
final prediction fusion. This identity concerns the aggregation step
and is not a statement about the total retraining dynamics or
generalization.
Let
$\mathcal{L}_i^{(t-1)}
=
\mathcal{D}_L\cup\mathcal{P}_i^{(t-1)}$
denote the augmented state associated with regressor
$\mathcal{M}_i$. Candidate admission is model-indexed because the
same local gain functional is evaluated separately on each
$\mathcal{L}_i^{(t-1)}$. Following the local error-reduction
principle of COREG, we estimate the local mean squared error
reduction resulting from accepting $(\mathbf{x}_u,\bar{y}_u)$ by 
\begin{equation}
\delta_{\mathbf{x}_u}^{(i)} =
\frac{1}{|\Omega_u|}
\sum_{\mathbf{x}_j\in\Omega_u}
\left[
\left(y_j - h_{i}^{\mathrm{temp}}(\mathbf{x}_j)\right)^2
-
\left(y_j - {h_{i}^{\mathrm{temp}}}^{\prime}(\mathbf{x}_j)\right)^2
\right],
\label{eq:guard2}
\end{equation}
where $h_i^{\mathrm{temp}}$ denotes the temporary $k$-NN proxy, common in form across regressors,
instantiated on $\mathcal{L}_i^{(t-1)}$,
$h_i^{\mathrm{temp}\prime}$ denotes the same proxy after temporarily
adding $(\mathbf{x}_u,\bar{y}_u)$, and $\Omega_u$ denotes the local
neighborhood of $\mathbf{x}_u$. The regressor index refers to the
augmented state on which the proxy is evaluated rather than to a
different proxy model class.

Because $\delta_{\mathbf{x}_u}^{(i)}$ is the difference between the
pre- and post-inclusion neighborhood MSE, any candidate satisfying
$\delta_{\mathbf{x}_u}^{(i)}>\theta$ lowers the neighborhood MSE of
the temporary proxy by more than $\theta$ when evaluated
individually. This local implication does not establish monotone
reduction of the global loss or test error after batch updating.

For each regressor $\mathcal{M}_i$, candidates satisfying
$\delta_{\mathbf{x}_u}^{(i)}>\theta$ form the feasible set
\begin{equation}
\mathcal{C}_i^{(t)}
=
\left\{
(\mathbf{x}_u,\bar{y}_u):
\mathbf{x}_u\in\mathcal{U}'_{t-1},\ 
\delta_{\mathbf{x}_u}^{(i)}>\theta
\right\}.
\end{equation}
CoGameReg then selects the top $\rho$ proportion of
$\mathcal{C}_i^{(t)}$, ranked by $\delta_{\mathbf{x}_u}^{(i)}$, to form the
accepted subset $\mathcal{A}_i^{(t)}$, and updates the pseudo-label pool by
\begin{equation}
\mathcal{P}_i^{(t)}=\mathcal{P}_i^{(t-1)}\cup \mathcal{A}_i^{(t)}.
\end{equation}
After all regressors complete their model-specific selections, the candidate pool is
updated as
\begin{equation}
\mathcal{U}'_{t}
=
\mathcal{U}'_{t-1}\setminus
\bigcup_{i=1}^{N}
\left\{
\mathbf{x}_u :
(\mathbf{x}_u,\bar{y}_u)\in \mathcal{A}_i^{(t)}
\right\}.
\end{equation}
Each regressor maintains its own augmented training set
$D_L \cup P_i^{(t)}$ and is retrained independently.
The heterogeneous regressors learn from the shared supervision
through their respective inductive biases.
The ensemble weights remain fixed throughout collaborative
training and are reused for final prediction fusion, i.e.,
$w^{*}=w^{(0)}$.
At each iteration, the shared pseudo-labels are recomputed
using these fixed weights and the current regressor predictions.

\section{Experimental Results and Analysis}
We evaluate CoGameReg in three complementary settings to examine its use for reservoir property prediction under sparse well control. Experiments on 10 public regression datasets compare predictive performance across datasets and labeled-data ratios, while ablations examine ensemble fusion, contribution-based weighting, and collaborative training. These benchmarks provide methodological evidence for the semi-supervised regression framework.
The two seismic experiments examine different prediction targets and input attributes. The synthetic experiment predicts P-wave velocity, an elastic property, from five attributes derived from post-stack seismic data. The field experiment predicts water saturation, a petrophysical property, from the post-stack seismic volume and inversion-derived impedance and velocity-ratio volumes. Both tasks combine sparse well labels with unlabeled attribute samples using the same learning framework; models are trained separately with their respective inputs, targets, and configurations. The synthetic experiment uses quantitative blind-well evaluation, qualitative slice comparisons against the known reference volume, and sensitivity analysis of neighborhood and weighting parameters. The field experiment compares four methods at common blind wells and on the same seismic section and reservoir horizon. The field volume comparisons provide spatial context, with saturation reference values available only along well trajectories.

\subsection{Benchmark Experiments on Public Datasets}

\subsubsection{Experimental Protocol}
Table~\ref{tab:datasets_info} summarizes the 10 public benchmark
regression datasets used in this study. The exact benchmark files were
selected from the data package distributed with the public
implementation of S2RMS \cite{ref_S2RMS}. The underlying datasets
originate from the UCI Machine Learning Repository~\cite{uci_repo}, Delve~\cite{delve_repo}, and StatLib~\cite{statlib_repo}. Following a unified protocol, we randomly select 2,000 samples from each dataset to form the training pool and use the remaining samples as the test set. The training pool is further partitioned into labeled and unlabeled subsets according to a labeled-data ratio $r \in \{2.5\%, 5\%, 10\%\}$, defined with respect to the training pool, yielding 50, 100, and 200 labeled samples per run, respectively. 
Twenty percent of the labeled subset is reserved as an internal
validation set for methods that require early stopping or hyperparameter
selection. For most methods, the remaining 80\% constitutes the labeled
training set. CoGameReg uses only this training portion both to fit the
base regressors and to estimate the initial Tri-Game Shapley weights; the
validation split is not used for weight estimation or collaborative
training. For TabM, early stopping on the validation subset determines
the training duration. The model is then reinitialized and refit from
scratch on the combined training and validation subsets for the selected
number of epochs. The CoGameReg weights are kept fixed throughout
collaborative training and final prediction. All methods are evaluated
over 10 independent runs using the same random seeds and data splits.
\begin{table}[!htbp]
      \centering
      \caption{Summary of the 10 public benchmark regression datasets.}
      \label{tab:datasets_info}

      \begin{tabular}{llllc}
          \toprule
          Name & Instances & Features & Target & Original source \\
          \midrule
          Abalone & 4,176 & 8 & Rings & UCI \\
          Bank32nh & 8,192 & 32 & rej & Delve \\
          Elevators & 9,517 & 6 & Se & Delve \\
          Folds5x2\_pp & 9,568 & 4 & PE & UCI \\
          Kin8nm & 8,192 & 8 & y & Delve \\
          Parkinsons & 5,875 & 21 & PPE & UCI \\
          Puma8NH & 8,192 & 8 & thetadd3 & Delve \\
          Space\_ga & 3,107 & 6 & YCOORD & StatLib \\
          Wind & 6,574 & 14 & MAL & StatLib \\
          Wine\_quality & 6,497 & 11 & quality & UCI \\
          \bottomrule
      \end{tabular}
  \end{table}

We compare CoGameReg against baselines spanning supervised, ensemble, self-training, classical semi-supervised, and recent neural and tabular paradigms. Specifically, the compared methods include
XGBoost, KNN, TabM~\cite{gorishniy2025tabm}, Avg.\ Ensemble, Self-KNN,
COREG~\cite{ref35}, Mean Teacher~\cite{tarvainen2017meanteacher},
SSDKL~\cite{jean2018ssdk}, and UCVME~\cite{dai2023ucvme}. COREG serves as the
most direct methodological reference because it is the classical co-training
baseline for regression, whereas Mean Teacher, SSDKL, and UCVME represent stronger
modern semi-supervised baselines. XGBoost and TabM provide competitive
supervised and tabular-learning references. 

For the benchmark datasets, performance is evaluated primarily in terms of
RMSE. To complement the per-dataset comparisons, we conduct Wilcoxon
signed-rank tests across the 10 datasets for each labeled-data ratio, together
with Holm correction and effect-size analysis. Detailed definitions of the
evaluation metrics and the statistical testing protocol are provided in
Appendix~B.1. The implementation details and fixed hyperparameter settings of
all baseline methods and base regressors are reported in Appendix~B.2.

\subsubsection{Comparison with Other Methods}

Table~\ref{tab:main_comparison} reports the RMSE results of all compared
methods on the 10 benchmark datasets under three labeled-data ratios.
Overall, CoGameReg achieves the lowest average RMSE ($0.1098$) across the
30 dataset-ratio settings and yields the lowest or joint-lowest RMSE in
18 of them. Specifically, it yields the lowest or joint-lowest RMSE on
7, 6, and 5 of the 10 datasets at the 2.5\%, 5\%, and 10\% labeled-data
ratios, respectively.

\begin{table}[!htbp]
\centering
\caption{Comparison of RMSE performance across different methods under varying labeled-data ratios.}
\label{tab:main_comparison}
\resizebox{\textwidth}{!}{
\begin{tabular}{ll*{10}{c}}
\toprule
Ratio & Dataset & XGBoost & KNN & TabM & Avg. Ensemble & Self-KNN & COREG & Mean Teacher & SSDKL & UCVME & CoGameReg \\
\midrule

\multirow{10}{*}{2.5\%}
& Abalone       & $0.1014_{\pm 0.0058}$ & $0.0979_{\pm 0.0021}$ & $0.0996_{\pm 0.0170}$ & $0.0927_{\pm 0.0062}$ & $0.0975_{\pm 0.0022}$ & $0.0971_{\pm 0.0030}$ & $\mathbf{0.0904_{\pm 0.0037}}$ & $0.1029_{\pm 0.0122}$ & $0.0978_{\pm 0.0068}$ & $\mathbf{0.0904_{\pm 0.0042}}$ \\
& Bank32nh      & $0.1415_{\pm 0.0075}$ & $0.1485_{\pm 0.0042}$ & $\mathbf{0.1316_{\pm 0.0204}}$ & $0.1478_{\pm 0.0146}$ & $0.1502_{\pm 0.0046}$ & $0.1478_{\pm 0.0035}$ & $0.1387_{\pm 0.0055}$ & $0.1451_{\pm 0.0065}$ & $0.1586_{\pm 0.0257}$ & $0.1342_{\pm 0.0071}$ \\
& Elevators     & $0.0670_{\pm 0.0040}$ & $0.0663_{\pm 0.0027}$ & $0.0841_{\pm 0.0247}$ & $0.0635_{\pm 0.0033}$ & $0.0646_{\pm 0.0024}$ & $0.0640_{\pm 0.0022}$ & $0.0630_{\pm 0.0027}$ & $0.0659_{\pm 0.0064}$ & $0.0674_{\pm 0.0070}$ & $\mathbf{0.0608_{\pm 0.0039}}$ \\
& Folds5x2\_pp  & $0.0746_{\pm 0.0056}$ & $0.0867_{\pm 0.0043}$ & $0.0736_{\pm 0.0171}$ & $0.0698_{\pm 0.0054}$ & $0.0855_{\pm 0.0049}$ & $0.0811_{\pm 0.0045}$ & $0.0706_{\pm 0.0063}$ & $0.0730_{\pm 0.0101}$ & $0.0883_{\pm 0.0135}$ & $\mathbf{0.0654_{\pm 0.0047}}$ \\
& Kin8nm        & $0.1626_{\pm 0.0055}$ & $0.1641_{\pm 0.0059}$ & $0.1682_{\pm 0.0194}$ & $0.1599_{\pm 0.0050}$ & $0.1638_{\pm 0.0051}$ & $0.1621_{\pm 0.0044}$ & $\mathbf{0.1514_{\pm 0.0040}}$ & $0.1576_{\pm 0.0108}$ & $0.1716_{\pm 0.0183}$ & $0.1539_{\pm 0.0046}$ \\
& Parkinsons    & $0.0759_{\pm 0.0090}$ & $0.0867_{\pm 0.0069}$ & $0.1270_{\pm 0.0479}$ & $0.1041_{\pm 0.0288}$ & $0.0884_{\pm 0.0068}$ & $0.0834_{\pm 0.0073}$ & $0.0783_{\pm 0.0080}$ & $0.0830_{\pm 0.0184}$ & $0.1140_{\pm 0.0390}$ & $\mathbf{0.0713_{\pm 0.0057}}$ \\
& Puma8NH       & $0.1889_{\pm 0.0090}$ & $0.2132_{\pm 0.0085}$ & $0.2320_{\pm 0.0493}$ & $0.1949_{\pm 0.0096}$ & $0.2075_{\pm 0.0070}$ & $0.2083_{\pm 0.0068}$ & $0.1998_{\pm 0.0165}$ & $0.2049_{\pm 0.0119}$ & $0.2231_{\pm 0.0223}$ & $\mathbf{0.1887_{\pm 0.0083}}$ \\
& Space\_ga     & $0.1852_{\pm 0.0092}$ & $0.1889_{\pm 0.0079}$ & $0.1925_{\pm 0.0348}$ & $\mathbf{0.1635_{\pm 0.0046}}$ & $0.1842_{\pm 0.0079}$ & $0.1865_{\pm 0.0084}$ & $0.1711_{\pm 0.0095}$ & $0.1868_{\pm 0.0354}$ & $0.1939_{\pm 0.0347}$ & $0.1650_{\pm 0.0058}$ \\
& Wind          & $0.1040_{\pm 0.0028}$ & $0.1117_{\pm 0.0037}$ & $0.1044_{\pm 0.0189}$ & $0.0970_{\pm 0.0032}$ & $0.1095_{\pm 0.0048}$ & $0.1079_{\pm 0.0032}$ & $0.0995_{\pm 0.0053}$ & $0.1035_{\pm 0.0065}$ & $0.1085_{\pm 0.0195}$ & $\mathbf{0.0934_{\pm 0.0040}}$ \\
& Wine\_quality & $0.1419_{\pm 0.0070}$ & $0.1461_{\pm 0.0037}$ & $0.1659_{\pm 0.0157}$ & $0.1401_{\pm 0.0058}$ & $0.1421_{\pm 0.0039}$ & $0.1436_{\pm 0.0037}$ & $0.1391_{\pm 0.0038}$ & $0.1520_{\pm 0.0197}$ & $0.1555_{\pm 0.0163}$ & $\mathbf{0.1348_{\pm 0.0049}}$ \\

\midrule

\multirow{10}{*}{5\%}
& Abalone       & $0.0921_{\pm 0.0036}$ & $0.0960_{\pm 0.0031}$ & $0.0931_{\pm 0.0065}$ & $0.0878_{\pm 0.0033}$ & $0.0963_{\pm 0.0030}$ & $0.0954_{\pm 0.0032}$ & $0.0875_{\pm 0.0022}$ & $0.1018_{\pm 0.0125}$ & $0.0902_{\pm 0.0033}$ & $\mathbf{0.0851_{\pm 0.0020}}$ \\
& Bank32nh      & $0.1329_{\pm 0.0071}$ & $0.1461_{\pm 0.0044}$ & $\mathbf{0.1160_{\pm 0.0114}}$ & $0.1269_{\pm 0.0052}$ & $0.1495_{\pm 0.0044}$ & $0.1459_{\pm 0.0045}$ & $0.1279_{\pm 0.0070}$ & $0.1375_{\pm 0.0080}$ & $0.1377_{\pm 0.0051}$ & $0.1249_{\pm 0.0040}$ \\
& Elevators     & $0.0629_{\pm 0.0021}$ & $0.0647_{\pm 0.0023}$ & $0.0615_{\pm 0.0054}$ & $0.0616_{\pm 0.0018}$ & $0.0628_{\pm 0.0021}$ & $0.0631_{\pm 0.0021}$ & $0.0594_{\pm 0.0019}$ & $0.0608_{\pm 0.0041}$ & $0.0599_{\pm 0.0019}$ & $\mathbf{0.0576_{\pm 0.0015}}$ \\
& Folds5x2\_pp  & $0.0677_{\pm 0.0039}$ & $0.0791_{\pm 0.0022}$ & $0.0652_{\pm 0.0065}$ & $0.0647_{\pm 0.0016}$ & $0.0785_{\pm 0.0019}$ & $0.0765_{\pm 0.0020}$ & $0.0656_{\pm 0.0020}$ & $0.0640_{\pm 0.0024}$ & $0.0722_{\pm 0.0072}$ & $\mathbf{0.0614_{\pm 0.0018}}$ \\
& Kin8nm        & $0.1563_{\pm 0.0019}$ & $0.1544_{\pm 0.0047}$ & $0.1371_{\pm 0.0113}$ & $0.1532_{\pm 0.0029}$ & $0.1559_{\pm 0.0040}$ & $0.1518_{\pm 0.0041}$ & $\mathbf{0.1369_{\pm 0.0061}}$ & $0.1452_{\pm 0.0070}$ & $0.1531_{\pm 0.0056}$ & $0.1467_{\pm 0.0025}$ \\
& Parkinsons    & $0.0680_{\pm 0.0038}$ & $0.0816_{\pm 0.0043}$ & $0.0894_{\pm 0.0278}$ & $0.0873_{\pm 0.0234}$ & $0.0841_{\pm 0.0052}$ & $0.0786_{\pm 0.0038}$ & $0.0716_{\pm 0.0038}$ & $0.0762_{\pm 0.0059}$ & $0.0903_{\pm 0.0192}$ & $\mathbf{0.0676_{\pm 0.0053}}$ \\
& Puma8NH       & $\mathbf{0.1749_{\pm 0.0068}}$ & $0.2086_{\pm 0.0069}$ & $0.1989_{\pm 0.0284}$ & $0.1832_{\pm 0.0045}$ & $0.2025_{\pm 0.0054}$ & $0.2028_{\pm 0.0065}$ & $0.1864_{\pm 0.0059}$ & $0.2045_{\pm 0.0066}$ & $0.2014_{\pm 0.0050}$ & $0.1770_{\pm 0.0048}$ \\
& Space\_ga     & $0.1708_{\pm 0.0027}$ & $0.1821_{\pm 0.0038}$ & $0.1628_{\pm 0.0047}$ & $\mathbf{0.1558_{\pm 0.0047}}$ & $0.1771_{\pm 0.0038}$ & $0.1802_{\pm 0.0025}$ & $0.1577_{\pm 0.0058}$ & $0.1689_{\pm 0.0104}$ & $0.1714_{\pm 0.0088}$ & $0.1565_{\pm 0.0054}$ \\
& Wind          & $0.0947_{\pm 0.0032}$ & $0.1050_{\pm 0.0027}$ & $0.0893_{\pm 0.0050}$ & $0.0923_{\pm 0.0036}$ & $0.1042_{\pm 0.0020}$ & $0.1019_{\pm 0.0030}$ & $0.0918_{\pm 0.0027}$ & $0.0921_{\pm 0.0029}$ & $0.0978_{\pm 0.0058}$ & $\mathbf{0.0883_{\pm 0.0026}}$ \\
& Wine\_quality & $0.1355_{\pm 0.0024}$ & $0.1430_{\pm 0.0040}$ & $0.1457_{\pm 0.0138}$ & $0.1343_{\pm 0.0035}$ & $0.1391_{\pm 0.0039}$ & $0.1400_{\pm 0.0040}$ & $0.1325_{\pm 0.0048}$ & $0.1418_{\pm 0.0105}$ & $0.1414_{\pm 0.0167}$ & $\mathbf{0.1300_{\pm 0.0033}}$ \\

\midrule

\multirow{10}{*}{10\%}
& Abalone       & $0.0893_{\pm 0.0041}$ & $0.0932_{\pm 0.0032}$ & $0.0861_{\pm 0.0062}$ & $0.0846_{\pm 0.0027}$ & $0.0934_{\pm 0.0029}$ & $0.0923_{\pm 0.0033}$ & $0.0833_{\pm 0.0030}$ & $0.0930_{\pm 0.0217}$ & $0.0855_{\pm 0.0034}$ & $\mathbf{0.0815_{\pm 0.0022}}$ \\
& Bank32nh      & $0.1222_{\pm 0.0035}$ & $0.1426_{\pm 0.0031}$ & $\mathbf{0.1083_{\pm 0.0035}}$ & $0.1192_{\pm 0.0035}$ & $0.1481_{\pm 0.0027}$ & $0.1417_{\pm 0.0033}$ & $0.1202_{\pm 0.0050}$ & $0.1275_{\pm 0.0087}$ & $0.1294_{\pm 0.0055}$ & $0.1166_{\pm 0.0036}$ \\
& Elevators     & $0.0607_{\pm 0.0005}$ & $0.0631_{\pm 0.0009}$ & $0.0571_{\pm 0.0023}$ & $0.0597_{\pm 0.0013}$ & $0.0620_{\pm 0.0008}$ & $0.0619_{\pm 0.0008}$ & $0.0575_{\pm 0.0007}$ & $0.0576_{\pm 0.0043}$ & $0.0579_{\pm 0.0025}$ & $\mathbf{0.0557_{\pm 0.0006}}$ \\
& Folds5x2\_pp  & $0.0648_{\pm 0.0021}$ & $0.0754_{\pm 0.0016}$ & $0.0596_{\pm 0.0025}$ & $0.0629_{\pm 0.0012}$ & $0.0746_{\pm 0.0020}$ & $0.0729_{\pm 0.0014}$ & $0.0631_{\pm 0.0018}$ & $0.0599_{\pm 0.0012}$ & $0.0635_{\pm 0.0022}$ & $\mathbf{0.0594_{\pm 0.0009}}$ \\
& Kin8nm        & $0.1437_{\pm 0.0038}$ & $0.1414_{\pm 0.0033}$ & $\mathbf{0.1031_{\pm 0.0047}}$ & $0.1418_{\pm 0.0027}$ & $0.1463_{\pm 0.0031}$ & $0.1391_{\pm 0.0030}$ & $0.1158_{\pm 0.0047}$ & $0.1322_{\pm 0.0049}$ & $0.1381_{\pm 0.0085}$ & $0.1360_{\pm 0.0022}$ \\
& Parkinsons    & $0.0637_{\pm 0.0017}$ & $0.0769_{\pm 0.0023}$ & $0.0687_{\pm 0.0065}$ & $0.0735_{\pm 0.0094}$ & $0.0796_{\pm 0.0026}$ & $0.0740_{\pm 0.0021}$ & $0.0656_{\pm 0.0020}$ & $0.0699_{\pm 0.0041}$ & $0.0836_{\pm 0.0155}$ & $\mathbf{0.0626_{\pm 0.0012}}$ \\
& Puma8NH       & $0.1624_{\pm 0.0041}$ & $0.1996_{\pm 0.0038}$ & $\mathbf{0.1565_{\pm 0.0223}}$ & $0.1752_{\pm 0.0030}$ & $0.1968_{\pm 0.0033}$ & $0.1937_{\pm 0.0040}$ & $0.1703_{\pm 0.0055}$ & $0.1931_{\pm 0.0080}$ & $0.1902_{\pm 0.0060}$ & $0.1695_{\pm 0.0031}$ \\
& Space\_ga     & $0.1600_{\pm 0.0046}$ & $0.1717_{\pm 0.0053}$ & $0.1566_{\pm 0.0065}$ & $0.1516_{\pm 0.0046}$ & $0.1680_{\pm 0.0047}$ & $0.1695_{\pm 0.0050}$ & $\mathbf{0.1486_{\pm 0.0048}}$ & $0.1624_{\pm 0.0128}$ & $0.1615_{\pm 0.0064}$ & $0.1509_{\pm 0.0044}$ \\
& Wind          & $0.0901_{\pm 0.0027}$ & $0.1011_{\pm 0.0023}$ & $\mathbf{0.0829_{\pm 0.0039}}$ & $0.0895_{\pm 0.0021}$ & $0.1016_{\pm 0.0013}$ & $0.0986_{\pm 0.0019}$ & $0.0882_{\pm 0.0021}$ & $0.0891_{\pm 0.0023}$ & $0.0903_{\pm 0.0051}$ & $0.0851_{\pm 0.0017}$ \\
& Wine\_quality & $0.1306_{\pm 0.0031}$ & $0.1381_{\pm 0.0035}$ & $0.1297_{\pm 0.0052}$ & $0.1284_{\pm 0.0033}$ & $0.1354_{\pm 0.0030}$ & $0.1357_{\pm 0.0037}$ & $0.1266_{\pm 0.0032}$ & $0.1332_{\pm 0.0090}$ & $0.1294_{\pm 0.0028}$ & $\mathbf{0.1250_{\pm 0.0031}}$ \\

\midrule
\multicolumn{2}{l}{Average RMSE} & $0.1162$ & $0.1258$ & $0.1182$ & $0.1156$ & $0.1250$ & $0.1232$ & $0.1119$ & $0.1195$ & $0.1241$ & $\mathbf{0.1098}$ \\

\bottomrule
\end{tabular}
}
\end{table}

The comparison with classical collaborative and self-training baselines is
particularly informative. CoGameReg achieves lower RMSE than Self-KNN, COREG,
and Avg.\ Ensemble on most datasets and labeled-data ratios. These results are
consistent with an empirical benefit from extending dual-regressor co-training
to a multi-regressor collaborative setting. The improvement over Avg.\
Ensemble also indicates that simple prediction averaging alone does not
account for the observed gains; the ablation analysis below further evaluates
the contributions of collaboration and contribution-aware weighting.

Strong modern baselines remain competitive on several datasets. Mean Teacher
performs strongly on Kin8nm and Space\_ga. SSDKL remains competitive on
Elevators at the 5\% and 10\% labeled-data ratios, while TabM becomes
increasingly competitive as the labeled-data ratio increases. Nevertheless,
CoGameReg achieves the lowest or joint-lowest RMSE on Abalone, Elevators,
Folds5x2\_pp, Parkinsons, Wind, and Wine\_quality at multiple labeled-data
ratios. Overall, CoGameReg achieves competitive predictive performance across the evaluated semi-supervised regression tasks.

\subsubsection{Statistical Significance Analysis}
To further assess the statistical significance of the observed performance differences, we conduct Wilcoxon signed-rank tests on the RMSE results across
the 10 datasets at each labeled-data ratio, with Holm correction to
account for multiple comparisons. The results are summarized in
Table~\ref{tab:wilcoxon_holm}.

\begin{table}[!htbp]
\centering
\caption{Per-ratio Wilcoxon signed-rank test results with Holm correction.}
\label{tab:wilcoxon_holm}
\scriptsize
\resizebox{\columnwidth}{!}{
\begin{tabular}{llcccc}
\toprule
Ratio & Method & W/T/L & $p$ & Holm-$p$ & $r$ \\
\midrule
\multirow{9}{*}{2.5\%}
 & XGBoost       & 10/0/0 & $1.95\mathrm{E}{-}03$ & $1.76\mathrm{E}{-}02$ & 0.886 \\
 & KNN           & 10/0/0 & $1.95\mathrm{E}{-}03$ & $1.76\mathrm{E}{-}02$ & 0.886 \\
 & TabM          & 9/0/1  & $3.91\mathrm{E}{-}03$ & $1.76\mathrm{E}{-}02$ & 0.854 \\
 & Avg. Ensemble & 9/0/1  & $3.91\mathrm{E}{-}03$ & $1.76\mathrm{E}{-}02$ & 0.854 \\
 & Self-KNN      & 10/0/0 & $1.95\mathrm{E}{-}03$ & $1.76\mathrm{E}{-}02$ & 0.886 \\
 & COREG         & 10/0/0 & $1.95\mathrm{E}{-}03$ & $1.76\mathrm{E}{-}02$ & 0.886 \\
 & Mean Teacher  & 8/1/1  & $1.17\mathrm{E}{-}02$ & $1.76\mathrm{E}{-}02$ & 0.810 \\
 & SSDKL         & 10/0/0 & $1.95\mathrm{E}{-}03$ & $1.76\mathrm{E}{-}02$ & 0.886 \\
 & UCVME         & 10/0/0 & $1.95\mathrm{E}{-}03$ & $1.76\mathrm{E}{-}02$ & 0.886 \\
\midrule
\multirow{9}{*}{5\%}
 & XGBoost       & 9/0/1  & $5.86\mathrm{E}{-}03$ & $1.95\mathrm{E}{-}02$ & 0.822 \\
 & KNN           & 10/0/0 & $1.95\mathrm{E}{-}03$ & $1.76\mathrm{E}{-}02$ & 0.886 \\
 & TabM          & 8/0/2  & $1.60\mathrm{E}{-}01$ & $1.68\mathrm{E}{-}01$ & 0.467 \\
 & Avg. Ensemble & 9/0/1  & $3.91\mathrm{E}{-}03$ & $1.95\mathrm{E}{-}02$ & 0.854 \\
 & Self-KNN      & 10/0/0 & $1.95\mathrm{E}{-}03$ & $1.76\mathrm{E}{-}02$ & 0.886 \\
 & COREG         & 10/0/0 & $1.95\mathrm{E}{-}03$ & $1.76\mathrm{E}{-}02$ & 0.886 \\
 & Mean Teacher  & 9/0/1  & $8.40\mathrm{E}{-}02$ & $1.68\mathrm{E}{-}01$ & 0.564 \\
 & SSDKL         & 9/0/1  & $3.91\mathrm{E}{-}03$ & $1.95\mathrm{E}{-}02$ & 0.854 \\
 & UCVME         & 10/0/0 & $1.95\mathrm{E}{-}03$ & $1.76\mathrm{E}{-}02$ & 0.886 \\
\midrule
\multirow{9}{*}{10\%}
 & XGBoost       & 9/0/1  & $3.71\mathrm{E}{-}02$ & $1.11\mathrm{E}{-}01$ & 0.661 \\
 & KNN           & 10/0/0 & $1.95\mathrm{E}{-}03$ & $1.76\mathrm{E}{-}02$ & 0.886 \\
 & TabM          & 6/0/4  & $8.46\mathrm{E}{-}01$ & $8.46\mathrm{E}{-}01$ & 0.081 \\
 & Avg. Ensemble & 10/0/0 & $1.95\mathrm{E}{-}03$ & $1.76\mathrm{E}{-}02$ & 0.886 \\
 & Self-KNN      & 10/0/0 & $1.95\mathrm{E}{-}03$ & $1.76\mathrm{E}{-}02$ & 0.886 \\
 & COREG         & 10/0/0 & $1.95\mathrm{E}{-}03$ & $1.76\mathrm{E}{-}02$ & 0.886 \\
 & Mean Teacher  & 8/0/2  & $2.32\mathrm{E}{-}01$ & $4.65\mathrm{E}{-}01$ & 0.403 \\
 & SSDKL         & 9/0/1  & $9.77\mathrm{E}{-}03$ & $3.91\mathrm{E}{-}02$ & 0.790 \\
 & UCVME         & 10/0/0 & $1.95\mathrm{E}{-}03$ & $1.76\mathrm{E}{-}02$ & 0.886 \\
\bottomrule
\end{tabular}
}
\end{table}

At the 2.5\% labeled-data ratio, CoGameReg has significantly lower RMSE
than all nine compared methods after Holm correction. It achieves a
win/tie/loss record of 10/0/0 against XGBoost, KNN, Self-KNN, COREG,
SSDKL, and UCVME; all six comparisons have Holm-adjusted $p$-values of
$0.0176$ and large effect sizes ($r = 0.886$). The comparisons with TabM
and Avg.\ Ensemble are also significant, each with a win/tie/loss record
of 9/0/1 (Holm-$p = 0.0176$, $r = 0.854$). Against Mean Teacher,
CoGameReg obtains a win/tie/loss record of 8/1/1
(Holm-$p = 0.0176$, $r = 0.810$).

At the 5\% labeled-data ratio, CoGameReg remains statistically superior to XGBoost, KNN,
Avg.\ Ensemble, Self-KNN, COREG, SSDKL, and UCVME after Holm correction. The
differences relative to TabM and Mean Teacher, however, are not statistically significant after Holm correction. This suggests that as the amount of supervision
increases, the advantage of CoGameReg over the stronger modern baselines becomes smaller,
although it still retains statistically significant advantages over the majority of the compared methods.

At the 10\% labeled-data ratio, CoGameReg has significantly lower RMSE
than KNN, Avg.\ Ensemble, Self-KNN, COREG, SSDKL, and UCVME after Holm
correction. By contrast, the differences relative to XGBoost, TabM, and
Mean Teacher are not statistically significant after Holm correction.
Against SSDKL, CoGameReg obtains a win/tie/loss record of 9/0/1
(Holm-$p = 0.0391$, $r = 0.790$). Although CoGameReg records more wins
than losses against TabM (6/0/4), this comparison remains statistically
non-significant (Holm-$p = 0.846$, $r = 0.081$).

Overall, CoGameReg shows robust and statistically significant improvements over
classical co-training, self-training, and simple ensemble baselines. Against
recent supervised and semi-supervised alternatives, its advantage is
most pronounced under scarce supervision and becomes increasingly
dataset-dependent as the labeled-data ratio increases. This pattern is consistent
with the design motivation of CoGameReg, namely, to exploit model complementarity across multiple regressors under
scarce supervision.

\subsubsection{Ablation Study}
To isolate the contributions of the main design components of CoGameReg, we conduct an ablation study on the same 10 benchmark datasets under three labeled-data ratios. The ablation variants are defined as follows:
\begin{itemize}
    \item \textbf{Split-Label CoGameReg (SL-CoGameReg)}: a variant in which 80\%
    of the labeled training set is used to train the base regressors, while
    the remaining 20\% is held out to compute the Tri-Game Shapley weights.

    \item \textbf{Average Ensemble (Avg.\ Ensemble)}: a purely supervised
    heterogeneous ensemble with uniform averaging and no semi-supervised
    updating.

    \item \textbf{Shapley Weighted Ensemble (Shap.\ Ensemble)}: a purely
    supervised heterogeneous ensemble that uses Tri-Game Shapley weights but
    does not involve semi-supervised collaboration.

    \item \textbf{$\alpha$-Only Shapley Weighted Ensemble ($\alpha$-Shap.\ Ensemble)}:
    a supervised Tri-Game weighted ensemble that uses only the Rank component,
    i.e., $\alpha=1$, $\beta=0$, and $\gamma=0$.

    \item \textbf{$\beta$-Only Shapley Weighted Ensemble ($\beta$-Shap.\ Ensemble)}:
    a supervised Tri-Game weighted ensemble that uses only the MSE component,
    i.e., $\alpha=0$, $\beta=1$, and $\gamma=0$.

    \item \textbf{$\gamma$-Only Shapley Weighted Ensemble ($\gamma$-Shap.\ Ensemble)}:
    a supervised Tri-Game weighted ensemble that uses only the Consistency
    component, i.e., $\alpha=0$, $\beta=0$, and $\gamma=1$.

    \item \textbf{SSR-Average Ensemble (SSR-Avg.\ Ensemble)}: semi-supervised
    collaborative training with uniform ensemble weights throughout both
    training and prediction.

    \item \textbf{SSR-Decoupled Shapley Weighting (Decoupled CoGameReg)}:
    semi-supervised collaborative training with uniform weights during
    pseudo-label generation, followed by Shapley-based weighting only at the
    final prediction stage.

    \item \textbf{SSR-Undecoupled Shapley Weighting (CoGameReg)}: the final CoGameReg
    variant, in which the initial Tri-Game Shapley weights are computed from
    the labeled data and then kept fixed during both semi-supervised pseudo-label
    generation and final prediction.
\end{itemize}

The results in Table~\ref{tab:ablation_variants} show that the final CoGameReg
variant achieves the lowest average RMSE ($0.1098$) among all ablation
variants and attains the best or tied-best result in 12 of the 30
dataset-ratio settings. It outperforms Split-Label CoGameReg, Average Ensemble,
Shapley Weighted Ensemble, SSR-Average Ensemble, and SSR-Decoupled Shapley
Weighting in 29, 28, 24, 30, and 27 settings, respectively.

\begin{table}[!htbp]
\centering
\caption{Comparison of RMSE performance across different ablation variants of CoGameReg under varying labeled-data ratios.}
\label{tab:ablation_variants}
\scriptsize
\setlength{\tabcolsep}{2.2pt}
\resizebox{\textwidth}{!}{
\begin{tabular}{llccccccccc}
\toprule
Ratio & Dataset & SL-CoGameReg & Avg.\ Ensemble & Shap.\ Ensemble & $\alpha$-Shap.\ Ensemble & $\beta$-Shap.\ Ensemble & $\gamma$-Shap.\ Ensemble & SSR-Avg.\ Ensemble & Decoupled CoGameReg & CoGameReg \\
\midrule
\multirow{10}{*}{2.5\%}
& Abalone & $0.0962_{\pm 0.0091}$ & $0.0927_{\pm 0.0062}$ & $0.0925_{\pm 0.0051}$ & $0.0913_{\pm 0.0054}$ & $0.1086_{\pm 0.0089}$ & $0.0909_{\pm 0.0060}$ & $0.0943_{\pm 0.0056}$ & $0.0911_{\pm 0.0042}$ & $\mathbf{0.0904_{\pm 0.0042}}$ \\
& Bank32nh & $0.1442_{\pm 0.0049}$ & $0.1478_{\pm 0.0146}$ & $0.1422_{\pm 0.0119}$ & $0.1443_{\pm 0.0151}$ & $0.1543_{\pm 0.0216}$ & $0.1466_{\pm 0.0158}$ & $0.1445_{\pm 0.0079}$ & $0.1364_{\pm 0.0067}$ & $\mathbf{0.1342_{\pm 0.0071}}$ \\
& Elevators & $0.0740_{\pm 0.0207}$ & $0.0635_{\pm 0.0033}$ & $0.0611_{\pm 0.0039}$ & $0.0606_{\pm 0.0034}$ & $0.0737_{\pm 0.0077}$ & $\mathbf{0.0588_{\pm 0.0033}}$ & $0.0692_{\pm 0.0054}$ & $0.0626_{\pm 0.0039}$ & $0.0608_{\pm 0.0039}$ \\
& Folds5x2\_pp & $0.0835_{\pm 0.0073}$ & $0.0698_{\pm 0.0054}$ & $0.0661_{\pm 0.0047}$ & $0.0692_{\pm 0.0053}$ & $0.0776_{\pm 0.0088}$ & $0.0674_{\pm 0.0052}$ & $0.0726_{\pm 0.0081}$ & $0.0682_{\pm 0.0071}$ & $\mathbf{0.0654_{\pm 0.0047}}$ \\
& Kin8nm & $0.1693_{\pm 0.0135}$ & $0.1599_{\pm 0.0050}$ & $0.1573_{\pm 0.0057}$ & $0.1568_{\pm 0.0047}$ & $0.1780_{\pm 0.0138}$ & $0.1562_{\pm 0.0045}$ & $0.1541_{\pm 0.0051}$ & $\mathbf{0.1528_{\pm 0.0046}}$ & $0.1539_{\pm 0.0046}$ \\
& Parkinsons & $0.0889_{\pm 0.0113}$ & $0.1041_{\pm 0.0288}$ & $0.0733_{\pm 0.0061}$ & $0.0882_{\pm 0.0199}$ & $0.0814_{\pm 0.0092}$ & $0.0780_{\pm 0.0092}$ & $0.0920_{\pm 0.0225}$ & $\mathbf{0.0711_{\pm 0.0071}}$ & $0.0713_{\pm 0.0057}$ \\
& Puma8NH & $0.2128_{\pm 0.0064}$ & $0.1949_{\pm 0.0096}$ & $\mathbf{0.1886_{\pm 0.0093}}$ & $0.1922_{\pm 0.0088}$ & $0.1997_{\pm 0.0155}$ & $0.1948_{\pm 0.0102}$ & $0.1958_{\pm 0.0078}$ & $0.1901_{\pm 0.0076}$ & $0.1887_{\pm 0.0083}$ \\
& Space\_ga & $0.1796_{\pm 0.0138}$ & $0.1635_{\pm 0.0046}$ & $0.1664_{\pm 0.0058}$ & $0.1639_{\pm 0.0052}$ & $0.1933_{\pm 0.0113}$ & $\mathbf{0.1633_{\pm 0.0043}}$ & $0.1658_{\pm 0.0059}$ & $0.1662_{\pm 0.0052}$ & $0.1650_{\pm 0.0058}$ \\
& Wind & $0.0967_{\pm 0.0057}$ & $0.0970_{\pm 0.0032}$ & $0.0942_{\pm 0.0024}$ & $0.0950_{\pm 0.0032}$ & $0.1118_{\pm 0.0048}$ & $\mathbf{0.0919_{\pm 0.0026}}$ & $0.0978_{\pm 0.0063}$ & $0.0941_{\pm 0.0051}$ & $0.0934_{\pm 0.0040}$ \\
& Wine\_quality & $0.1407_{\pm 0.0061}$ & $0.1401_{\pm 0.0058}$ & $0.1384_{\pm 0.0059}$ & $0.1382_{\pm 0.0052}$ & $0.1571_{\pm 0.0124}$ & $0.1369_{\pm 0.0046}$ & $0.1406_{\pm 0.0056}$ & $0.1358_{\pm 0.0049}$ & $\mathbf{0.1348_{\pm 0.0049}}$ \\

\midrule

\multirow{10}{*}{5\%}
& Abalone & $0.0857_{\pm 0.0049}$ & $0.0878_{\pm 0.0033}$ & $0.0864_{\pm 0.0024}$ & $0.0862_{\pm 0.0027}$ & $0.0980_{\pm 0.0054}$ & $0.0859_{\pm 0.0029}$ & $0.0888_{\pm 0.0044}$ & $0.0860_{\pm 0.0029}$ & $\mathbf{0.0851_{\pm 0.0020}}$ \\
& Bank32nh & $0.1297_{\pm 0.0063}$ & $0.1269_{\pm 0.0052}$ & $0.1226_{\pm 0.0038}$ & $0.1216_{\pm 0.0043}$ & $0.1431_{\pm 0.0097}$ & $\mathbf{0.1208_{\pm 0.0042}}$ & $0.1355_{\pm 0.0060}$ & $0.1269_{\pm 0.0040}$ & $0.1249_{\pm 0.0040}$ \\
& Elevators & $0.0581_{\pm 0.0019}$ & $0.0616_{\pm 0.0018}$ & $0.0586_{\pm 0.0022}$ & $0.0582_{\pm 0.0018}$ & $0.0690_{\pm 0.0052}$ & $\mathbf{0.0567_{\pm 0.0016}}$ & $0.0651_{\pm 0.0033}$ & $0.0590_{\pm 0.0024}$ & $0.0576_{\pm 0.0015}$ \\
& Folds5x2\_pp & $0.0688_{\pm 0.0046}$ & $0.0647_{\pm 0.0016}$ & $0.0618_{\pm 0.0016}$ & $0.0643_{\pm 0.0016}$ & $0.0706_{\pm 0.0034}$ & $0.0625_{\pm 0.0015}$ & $0.0670_{\pm 0.0039}$ & $0.0627_{\pm 0.0025}$ & $\mathbf{0.0614_{\pm 0.0018}}$ \\
& Kin8nm & $0.1516_{\pm 0.0024}$ & $0.1532_{\pm 0.0029}$ & $0.1484_{\pm 0.0028}$ & $0.1477_{\pm 0.0027}$ & $0.1657_{\pm 0.0052}$ & $0.1473_{\pm 0.0027}$ & $0.1496_{\pm 0.0039}$ & $0.1474_{\pm 0.0026}$ & $\mathbf{0.1467_{\pm 0.0025}}$ \\
& Parkinsons & $0.0782_{\pm 0.0095}$ & $0.0873_{\pm 0.0234}$ & $0.0684_{\pm 0.0078}$ & $0.0763_{\pm 0.0177}$ & $0.0722_{\pm 0.0044}$ & $0.0734_{\pm 0.0147}$ & $0.0828_{\pm 0.0162}$ & $0.0679_{\pm 0.0032}$ & $\mathbf{0.0676_{\pm 0.0053}}$ \\
& Puma8NH & $0.1867_{\pm 0.0013}$ & $0.1832_{\pm 0.0045}$ & $\mathbf{0.1740_{\pm 0.0062}}$ & $0.1791_{\pm 0.0049}$ & $0.1805_{\pm 0.0099}$ & $0.1839_{\pm 0.0044}$ & $0.1874_{\pm 0.0061}$ & $0.1810_{\pm 0.0060}$ & $0.1770_{\pm 0.0048}$ \\
& Space\_ga & $0.1651_{\pm 0.0056}$ & $0.1558_{\pm 0.0047}$ & $0.1566_{\pm 0.0034}$ & $\mathbf{0.1554_{\pm 0.0042}}$ & $0.1773_{\pm 0.0065}$ & $0.1561_{\pm 0.0050}$ & $0.1584_{\pm 0.0062}$ & $0.1578_{\pm 0.0051}$ & $0.1565_{\pm 0.0054}$ \\
& Wind & $0.0937_{\pm 0.0021}$ & $0.0923_{\pm 0.0036}$ & $0.0891_{\pm 0.0034}$ & $0.0896_{\pm 0.0031}$ & $0.1032_{\pm 0.0063}$ & $\mathbf{0.0872_{\pm 0.0029}}$ & $0.0932_{\pm 0.0045}$ & $0.0893_{\pm 0.0033}$ & $0.0883_{\pm 0.0026}$ \\
& Wine\_quality & $0.1319_{\pm 0.0055}$ & $0.1343_{\pm 0.0035}$ & $0.1316_{\pm 0.0027}$ & $0.1313_{\pm 0.0031}$ & $0.1443_{\pm 0.0038}$ & $0.1304_{\pm 0.0035}$ & $0.1342_{\pm 0.0057}$ & $\mathbf{0.1299_{\pm 0.0038}}$ & $0.1300_{\pm 0.0033}$ \\

\midrule

\multirow{10}{*}{10\%}
& Abalone & $\mathbf{0.0815_{\pm 0.0015}}$ & $0.0846_{\pm 0.0027}$ & $0.0831_{\pm 0.0024}$ & $0.0831_{\pm 0.0024}$ & $0.0932_{\pm 0.0044}$ & $0.0825_{\pm 0.0023}$ & $0.0843_{\pm 0.0033}$ & $0.0822_{\pm 0.0022}$ & $\mathbf{0.0815_{\pm 0.0022}}$ \\
& Bank32nh & $0.1180_{\pm 0.0050}$ & $0.1192_{\pm 0.0035}$ & $0.1150_{\pm 0.0037}$ & $0.1142_{\pm 0.0034}$ & $0.1300_{\pm 0.0045}$ & $\mathbf{0.1130_{\pm 0.0035}}$ & $0.1257_{\pm 0.0042}$ & $0.1182_{\pm 0.0034}$ & $0.1166_{\pm 0.0036}$ \\
& Elevators & $0.0572_{\pm 0.0013}$ & $0.0597_{\pm 0.0013}$ & $0.0563_{\pm 0.0006}$ & $0.0563_{\pm 0.0010}$ & $0.0638_{\pm 0.0010}$ & $\mathbf{0.0552_{\pm 0.0005}}$ & $0.0606_{\pm 0.0012}$ & $0.0564_{\pm 0.0007}$ & $0.0557_{\pm 0.0006}$ \\
& Folds5x2\_pp & $0.0631_{\pm 0.0023}$ & $0.0629_{\pm 0.0012}$ & $0.0598_{\pm 0.0007}$ & $0.0625_{\pm 0.0011}$ & $0.0667_{\pm 0.0026}$ & $0.0605_{\pm 0.0009}$ & $0.0636_{\pm 0.0033}$ & $0.0608_{\pm 0.0022}$ & $\mathbf{0.0594_{\pm 0.0009}}$ \\
& Kin8nm & $0.1413_{\pm 0.0033}$ & $0.1418_{\pm 0.0027}$ & $0.1361_{\pm 0.0027}$ & $\mathbf{0.1353_{\pm 0.0027}}$ & $0.1532_{\pm 0.0039}$ & $0.1358_{\pm 0.0022}$ & $0.1398_{\pm 0.0031}$ & $0.1375_{\pm 0.0026}$ & $0.1360_{\pm 0.0022}$ \\
& Parkinsons & $0.0684_{\pm 0.0017}$ & $0.0735_{\pm 0.0094}$ & $\mathbf{0.0623_{\pm 0.0020}}$ & $0.0653_{\pm 0.0053}$ & $0.0684_{\pm 0.0020}$ & $0.0640_{\pm 0.0046}$ & $0.0705_{\pm 0.0015}$ & $0.0640_{\pm 0.0012}$ & $0.0626_{\pm 0.0012}$ \\
& Puma8NH & $0.1822_{\pm 0.0038}$ & $0.1752_{\pm 0.0030}$ & $\mathbf{0.1642_{\pm 0.0033}}$ & $0.1701_{\pm 0.0029}$ & $0.1650_{\pm 0.0045}$ & $0.1757_{\pm 0.0032}$ & $0.1804_{\pm 0.0030}$ & $0.1743_{\pm 0.0034}$ & $0.1695_{\pm 0.0031}$ \\
& Space\_ga & $0.1610_{\pm 0.0063}$ & $0.1516_{\pm 0.0046}$ & $0.1516_{\pm 0.0048}$ & $\mathbf{0.1509_{\pm 0.0047}}$ & $0.1688_{\pm 0.0058}$ & $0.1518_{\pm 0.0045}$ & $0.1523_{\pm 0.0039}$ & $0.1519_{\pm 0.0039}$ & $\mathbf{0.1509_{\pm 0.0044}}$ \\
& Wind & $0.0887_{\pm 0.0017}$ & $0.0895_{\pm 0.0021}$ & $0.0856_{\pm 0.0020}$ & $0.0863_{\pm 0.0017}$ & $0.0970_{\pm 0.0033}$ & $\mathbf{0.0839_{\pm 0.0016}}$ & $0.0885_{\pm 0.0024}$ & $0.0860_{\pm 0.0018}$ & $0.0851_{\pm 0.0017}$ \\
& Wine\_quality & $0.1273_{\pm 0.0013}$ & $0.1284_{\pm 0.0033}$ & $0.1263_{\pm 0.0029}$ & $0.1256_{\pm 0.0030}$ & $0.1379_{\pm 0.0031}$ & $\mathbf{0.1250_{\pm 0.0030}}$ & $0.1290_{\pm 0.0051}$ & $0.1257_{\pm 0.0040}$ & $\mathbf{0.1250_{\pm 0.0031}}$ \\

\midrule
\multicolumn{2}{l}{Average RMSE}
& $0.1175$
& $0.1156$
& $0.1106$
& $0.1120$
& $0.1234$
& $0.1112$
& $0.1161$
& $0.1111$
& $\mathbf{0.1098}$ \\
\bottomrule
\end{tabular}
}
\end{table}

First, contribution-aware fusion remains beneficial even without semi-supervised updating:
compared with Average Ensemble ($0.1156$), Shapley Weighted Ensemble reduces
the average RMSE to $0.1106$. This suggests that explicitly modeling regressor contributions is more effective than uniform averaging in the purely supervised setting.

Second, the single-component variants reveal that the three Tri-Game
components play distinct empirical roles. The $\gamma$-only variant achieves
a competitive average RMSE ($0.1112$), indicating that cross-model consensus
provides a useful signal for supervised multi-model fusion. The $\alpha$-only
variant is also competitive ($0.1120$), suggesting that ordinal information
provides a complementary signal. By contrast, the $\beta$-only variant performs
substantially worse ($0.1234$), showing that the MSE-based utility alone is
insufficient for achieving strong performance. Among the purely supervised
weighting variants, the full Shapley Weighted Ensemble achieves the lowest
average RMSE ($0.1106$), slightly outperforming the $\gamma$-only and
$\alpha$-only variants on average. Although the $\gamma$-only variant achieves
a lower RMSE than the full Shapley Weighted Ensemble in 19 of the 30 individual
settings, the full combination has the lower overall average RMSE. Thus, the
results support an average advantage for combining the three utility
components, but not a consistent advantage across individual settings. More
importantly, the final CoGameReg variant achieves a lower average RMSE
($0.1098$) than all single-component and purely supervised variants.

Third, the effect of semi-supervised collaboration depends on how the
contribution-aware weights are incorporated. SSR-Decoupled Shapley Weighting
achieves an average RMSE of $0.1111$, which is slightly higher than that of the
purely supervised Shapley Weighted Ensemble ($0.1106$). Therefore, decoupled
pseudo-label training does not provide an additional average gain over purely
supervised weighted fusion in this comparison. In contrast, the final
CoGameReg variant reduces the average RMSE to $0.1098$ and outperforms both.
This suggests that the benefit of semi-supervised collaboration is associated
with using the fixed contribution-aware weights during pseudo-label generation
and final prediction, rather than applying Shapley weights only at final
fusion.

Finally, among the collaborative CoGameReg variants, Split-Label CoGameReg has
the highest average RMSE ($0.1175$). This suggests that, when labeled data are
scarce, splitting the labeled set into separate subsets for base-regressor
training and contribution-weight estimation may be suboptimal. Such a split
reduces the effective amount of labeled data available to train each base
regressor, while the Shapley weights estimated from a smaller labeled subset
may also become less stable. This observation supports the final design choice
of initializing both the base regressors and the contribution-aware weights
using all available labeled data and then keeping these weights fixed
throughout collaborative training.

Overall, the ablation results suggest that the best average performance is
obtained when multi-model ensemble diversity, multi-component Tri-Game Shapley
valuation, and fixed-initial-weight semi-supervised collaborative training are
combined. No single-component or intermediate variant matches the average RMSE
of the complete CoGameReg configuration.

\begin{figure}[!htbp]
\centering
\setlength{\tabcolsep}{0.01\textwidth}

\newcommand{\synimg}[1]{%
  \includegraphics[
    width=0.31\textwidth,
    height=0.33\textwidth,
    keepaspectratio
  ]{figures/picture/synthetic_input/#1}%
}

\newcommand{\synlabel}[1]{%
  \parbox[t]{0.31\textwidth}{%
    \centering\small #1%
  }%
}

\begin{tabular}{@{}ccc@{}}

\synimg{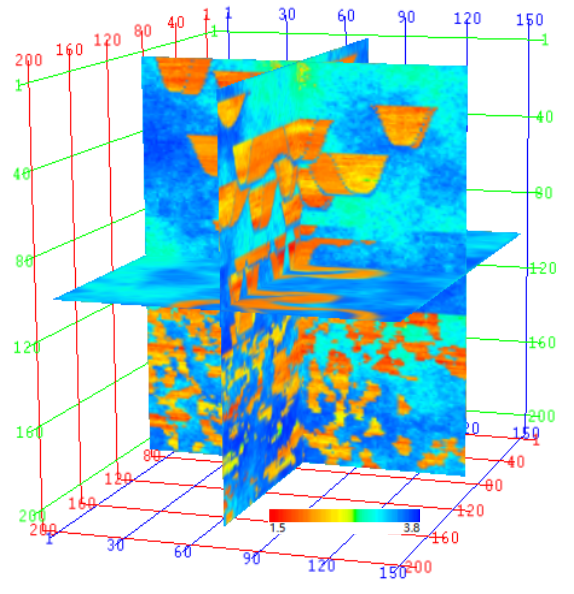}
&
\synimg{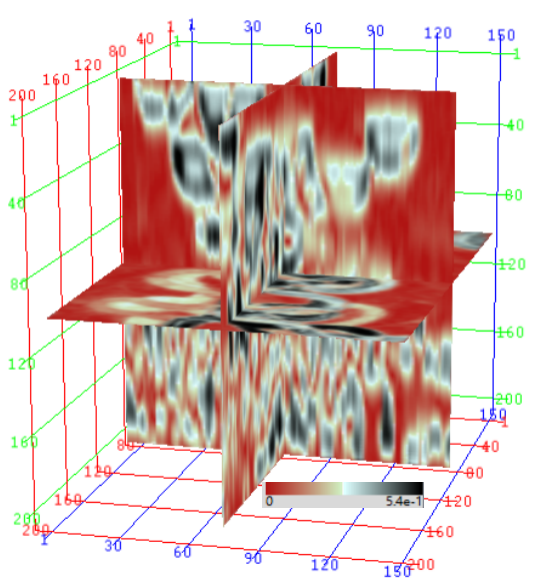}
&
\synimg{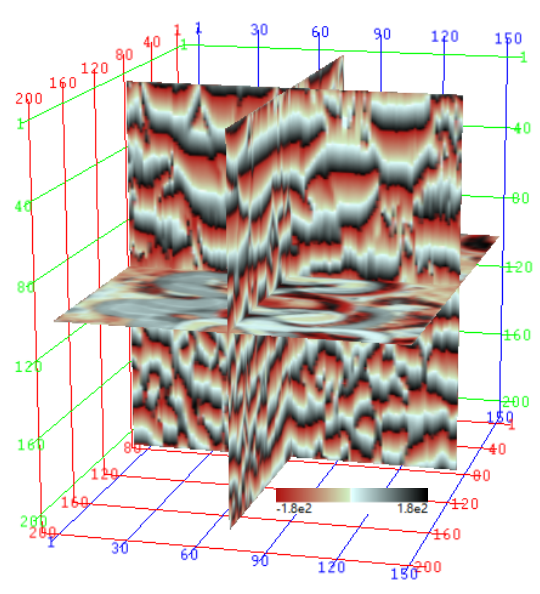}
\\[4pt]

\synlabel{(a) Ground-truth $V_p$}
&
\synlabel{(b) Instantaneous\\amplitude}
&
\synlabel{(c) Instantaneous\\phase}
\\[12pt]

\synimg{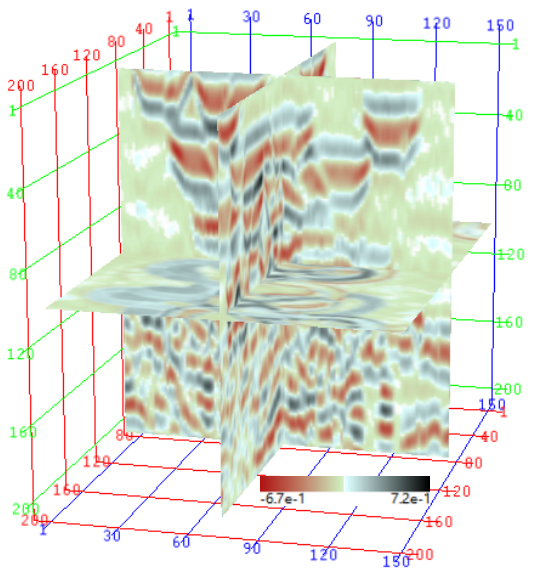}
&
\synimg{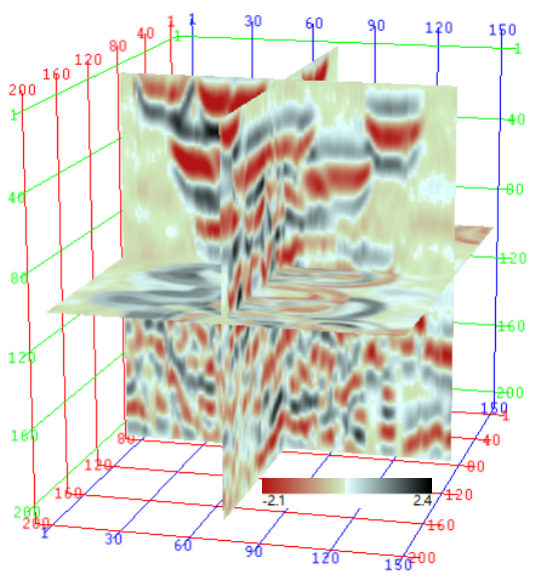}
&
\synimg{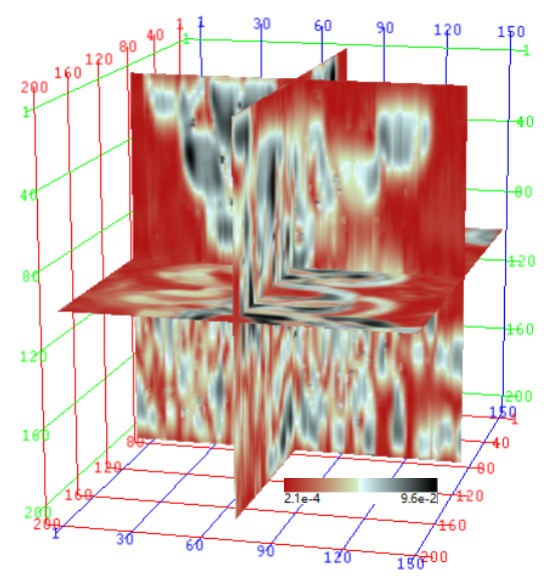}
\\[4pt]

\synlabel{(d) Instantaneous\\quadrature}
&
\synlabel{(e) Integral band}
&
\synlabel{(f) Sweetness}

\end{tabular}

\caption{Ground-truth P-wave velocity and five seismic
attributes used as inputs for the synthetic experiment.
Each panel is displayed with its own color scale.}
\label{fig:synthetic_inputs}
\end{figure}

\subsection{Synthetic 3D Seismic Reservoir Property Prediction}
\label{sec:simulated_3d}

\subsubsection{3D Seismic Data Configuration and Evaluation Metrics}

To investigate the applicability of CoGameReg to seismic reservoir characterization, we further conduct experiments on a simulated 3D seismic reservoir property prediction task in the petroleum exploration domain, for which full-field ground truth is available. This 3D setting enables us to assess local prediction quality at blind wells and to visually examine structural consistency on representative slices.

\begin{figure}[htbp]
\centering
\includegraphics[width=0.6\columnwidth]
{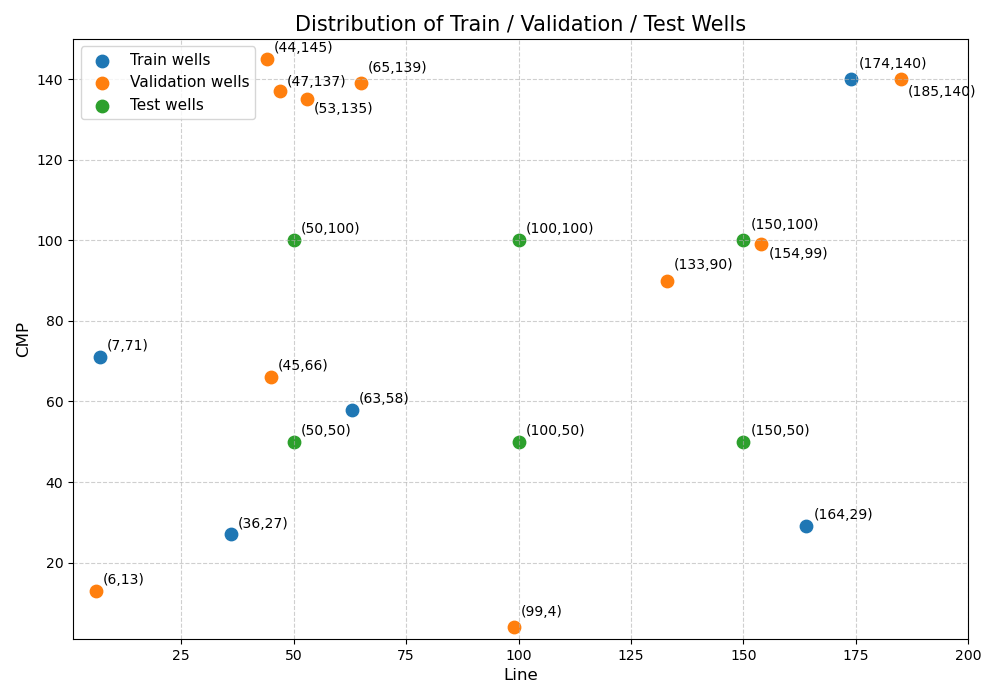}
\caption{Spatial distribution of the training, validation,
and blind-test wells in the synthetic experiment.}
\label{fig:synthetic_well_locations}
\end{figure}

We use the Stanford VI-E synthetic geological model~\cite{ref100} as the simulation benchmark. The dataset consists of a large-scale 3D reservoir volume with approximately $6\times10^6$ cells, with a grid size of $150\times200\times200$. The horizontal and vertical resolutions are $25\,\mathrm{m}$ and $1\,\mathrm{m}$, respectively. 
The prediction target is P-wave velocity, whose ground-truth
distribution is shown in Figure~\ref{fig:synthetic_inputs}(a).
The available wells are divided into training, validation,
and blind-test sets, with their spatial distribution shown
in Figure~\ref{fig:synthetic_well_locations}.
This well split is generated by random sampling subject to spatial coverage and separation constraints. Specifically, the blind test wells are selected to be spatially well distributed across the survey area and sufficiently separated from one another, so that the evaluation better reflects generalization at representative blind locations rather than interpolation among clustered wells. The training wells are used for model fitting, the validation wells for parameter selection and sensitivity analysis, and the blind test wells are reserved exclusively for final evaluation.
The five seismic attributes used as model inputs are displayed
in Figure~\ref{fig:synthetic_inputs}(b--f). The five seismic attribute volumes were derived from the post-stack seismic data using GeoEast and supplied as precomputed inputs for P-wave velocity prediction.
The present experiment evaluates how CoGameReg combines these
attributes with sparse well labels to predict the target property.

In this setting, quantitative evaluation is performed at blind wells using RMSE, $R^2$, and Spearman rank correlation (SRC). Because full-field ground truth is available, we also provide qualitative structural comparisons on representative slices to examine geological continuity and lateral variation. For the internal heterogeneous ensemble of CoGameReg, we use the same six representative base regressors as in the benchmark experiments: XGBoost, SVR, MLP, Ridge, Linear Regression, and KNN. The external baselines include XGBoost, KNN, TabM, Avg.\ Ensemble, Self-KNN, COREG, Mean Teacher, SSDKL, and UCVME.
A task-specific Tri-Game coefficient setting was selected using the
validation wells. The complete synthetic-data configuration and CoGameReg
parameter settings are provided in Appendix~B.2 of the supplementary
material, and the corresponding sensitivity analysis is reported in
Appendix~C.

\subsubsection{Blind-Well Evaluation and Qualitative Comparison}
Table~\ref{tab:main_results} reports the blind-well quantitative results for all
compared methods. Figure~\ref{fig:qualitative_multi_slices}
focuses on representative supervised and classical
semi-supervised regression approaches. From left to right,
the first two columns show XGBoost and KNN as supervised
references, followed by Self-KNN and COREG as self-training
and co-training baselines. CoGameReg is placed immediately
to the left of the ground truth to facilitate direct
comparison. Additional results for the remaining methods
are presented in Figure~S3 of Appendix~D.

\begin{table}[!htbp]
  \centering
  \caption{Quantitative comparison of reservoir property prediction on the simulated 3D dataset, evaluated on blind wells.}
  \label{tab:main_results}
  \begin{tabular}{lccc}
    \toprule
    \textbf{Method} & \textbf{RMSE $\downarrow$} & \textbf{$R^2$ $\uparrow$} & \textbf{SRC $\uparrow$} \\
    \midrule
    XGBoost            & 0.4580 & 0.3954 & 0.5245 \\
    KNN                & 0.4805 & 0.3345 & 0.5062 \\ 
    TabM               & 0.4230 & 0.4841 & 0.5686 \\   
    Avg. Ensemble      & 0.4149 & 0.5036 & 0.5732 \\
    Self-KNN           & 0.4713 & 0.3598 & 0.5183 \\
    COREG              & 0.4629 & 0.3822 & 0.5213 \\
    Mean Teacher       & 0.4188 & 0.4944 & 0.5663 \\
    SSDKL              & 0.4223 & 0.4858 & 0.5668 \\
    UCVME              & 0.4139 & 0.5061 & 0.5727 \\
    CoGameReg      & \textbf{0.4137} & \textbf{0.5065} & \textbf{0.5747} \\
    \bottomrule
  \end{tabular}
\end{table}

CoGameReg achieves an RMSE of 0.4137, an $R^2$ of 0.5065,
and an SRC of 0.5747, yielding the best values among the
methods reported in Table~5. Relative to XGBoost, KNN,
Self-KNN, and COREG, its RMSE is reduced by approximately
9.7\%, 13.9\%, 12.2\%, and 10.6\%, respectively. Both
Self-KNN and COREG improve upon the supervised KNN
baseline, whereas CoGameReg achieves lower prediction
error than all four reference methods. These comparisons
support the effectiveness of the complete collaborative
framework for blind-well prediction in this synthetic
experiment.

The first two columns of
Figure~\ref{fig:qualitative_multi_slices} show that XGBoost
and KNN recover the main spatial patterns, including the
curved bands at Time = 9, the meandering structures at
Time = 110, and the branching patterns at Time = 188.
However, local fluctuations remain visible within and
around the principal structures. The Self-KNN and COREG
predictions in the next two columns preserve similar
large-scale geometries, but background mottling and
irregular local variations are still evident. Their
quantitative improvements over KNN therefore do not
translate into uniformly clearer structures throughout
the displayed slices.

The CoGameReg predictions in the fifth column exhibit
less mottled background variation than the four reference
predictions, particularly at Time = 9 and Time = 110.
The background surrounding the bodies in the upper part
of the Inline = 121 section is also smoother, making
the principal structures more clearly distinguishable.
Comparison with the rightmost ground-truth column shows
that CoGameReg preserves the main channel patterns,
although narrow internal bands, smoothed boundaries,
and incomplete recovery of fine-scale heterogeneity
remain apparent. These limitations are particularly
visible in the Time = 188 slice and the lower part of
the inline section. The visual results thus complement
the blind-well evaluation by showing improved spatial
coherence.

Taken together, the results demonstrate an advantage
over the supervised and classical semi-supervised
references highlighted in the main figure, while
maintaining competitive performance relative to the
remaining methods.

\newcommand{\qualimg}[1]{%
    \includegraphics[width=0.145\textwidth]{#1}%
}
\newcommand{\rowlab}[1]{%
    \rotatebox{90}{\scriptsize #1}%
}

\begin{figure}[!htbp]
\centering
\setlength{\tabcolsep}{2pt}
\renewcommand{\arraystretch}{0.95}

\begin{tabular}{@{}c@{\hspace{3pt}}cccccc@{}}
\rowlab{Time = 9}
& \qualimg{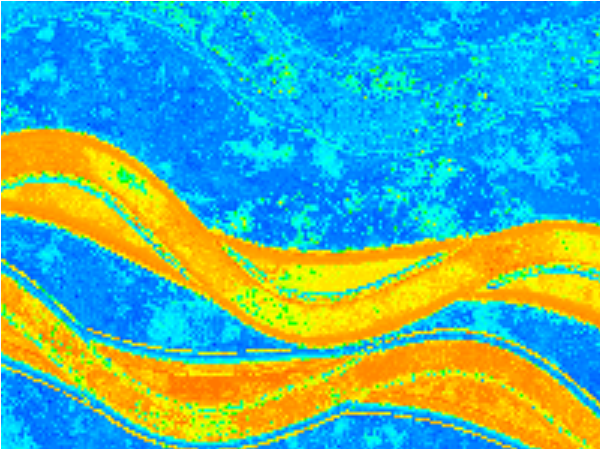}
& \qualimg{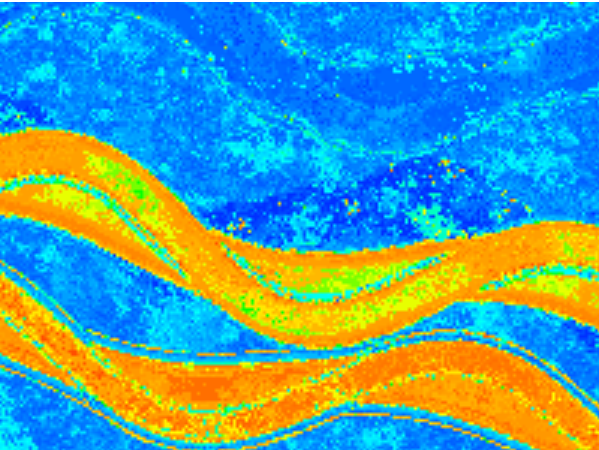}
& \qualimg{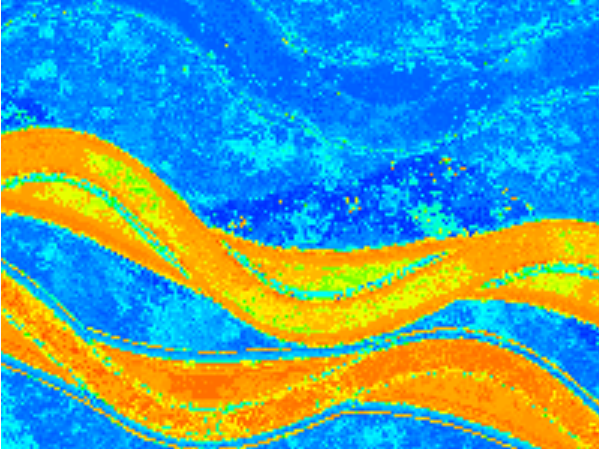}
& \qualimg{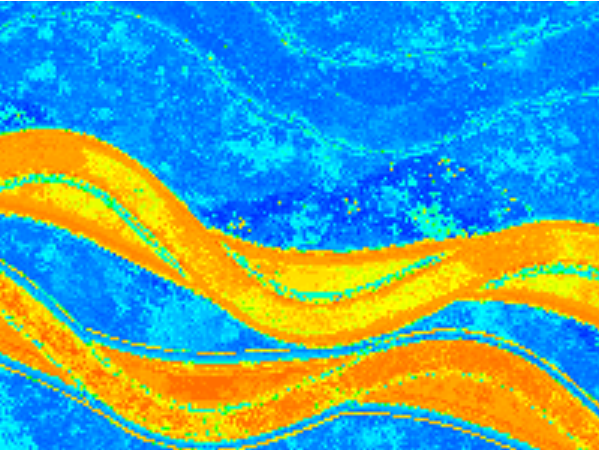}
& \qualimg{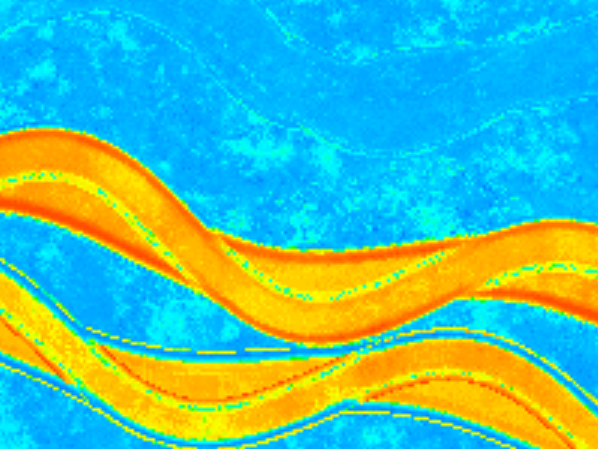}
& \qualimg{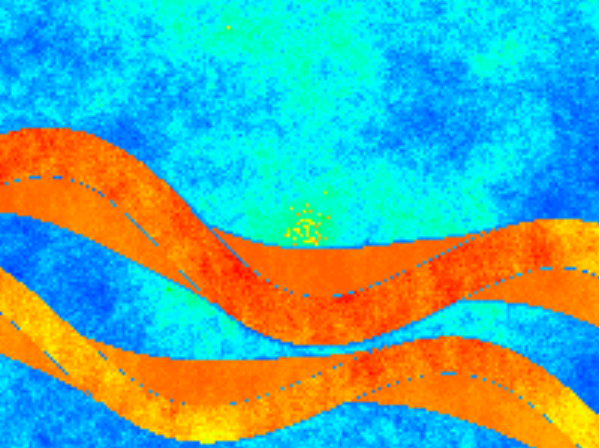}
\\[2pt]

\rowlab{Time = 110}
& \qualimg{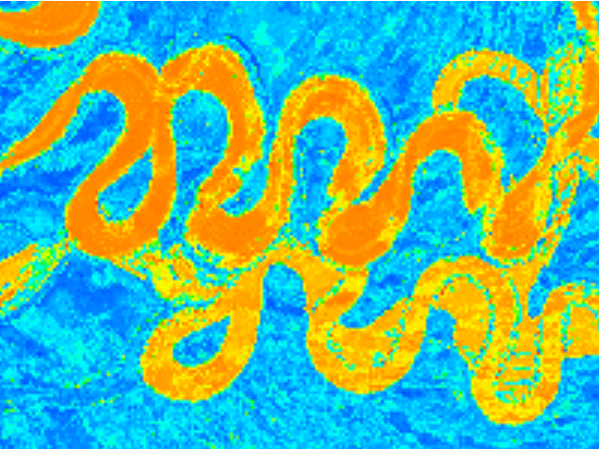}
& \qualimg{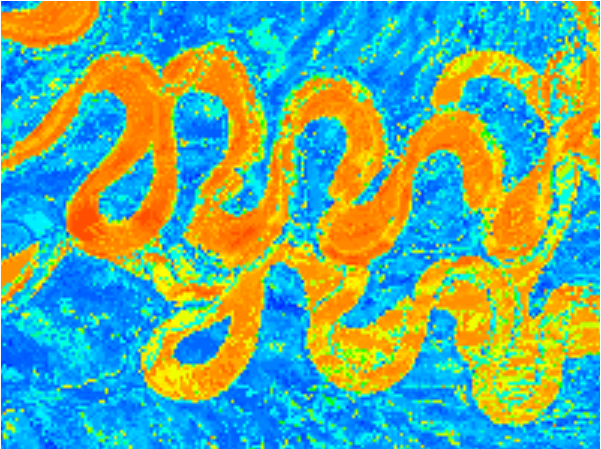}
& \qualimg{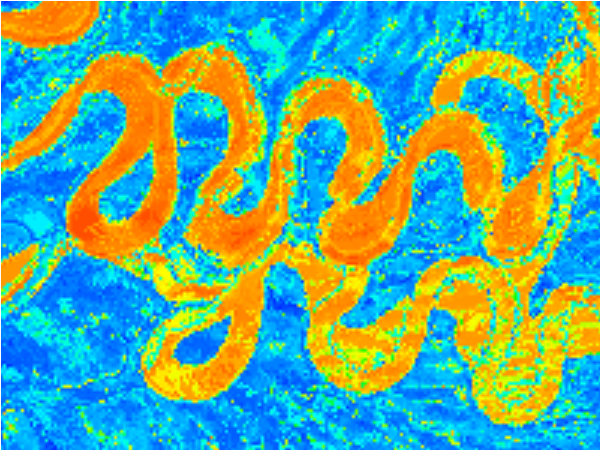}
& \qualimg{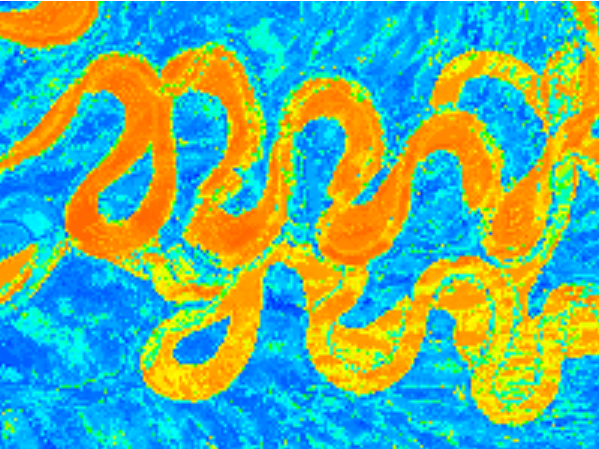}
& \qualimg{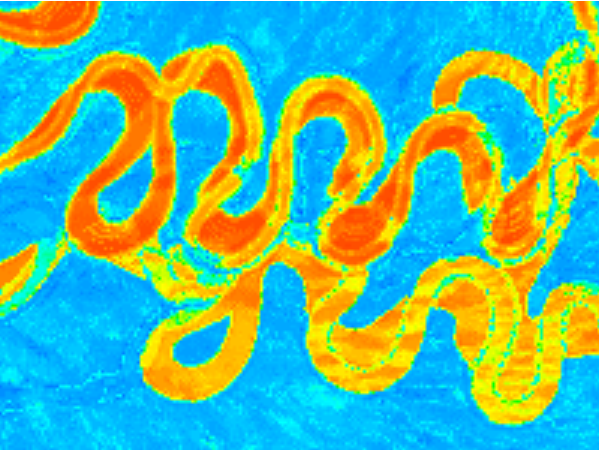}
& \qualimg{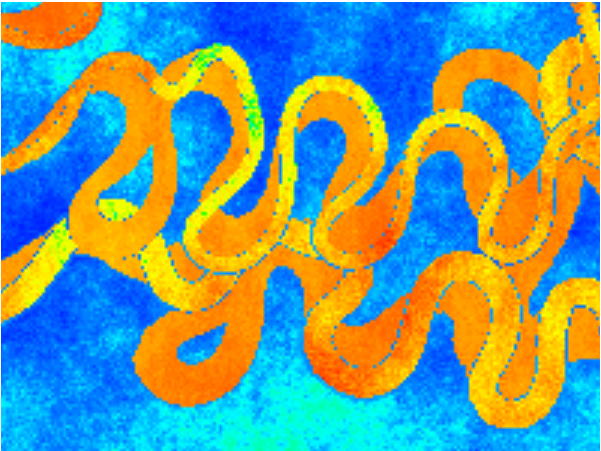}
\\[2pt]

\rowlab{Time = 188}
& \qualimg{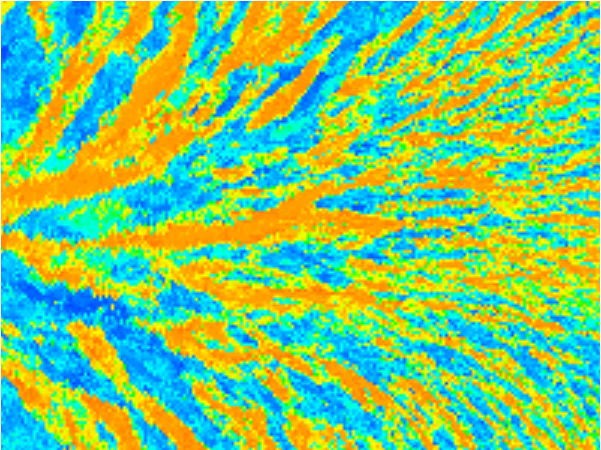}
& \qualimg{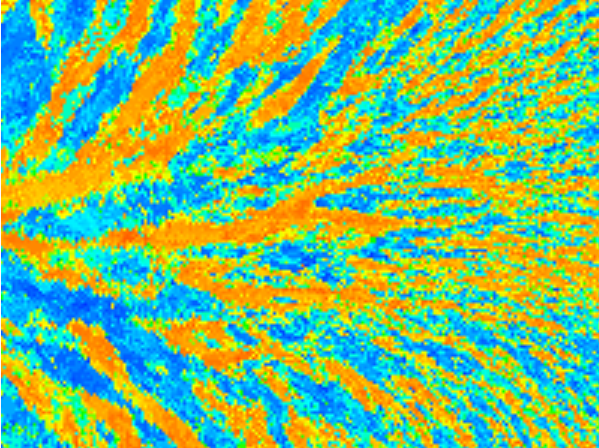}
& \qualimg{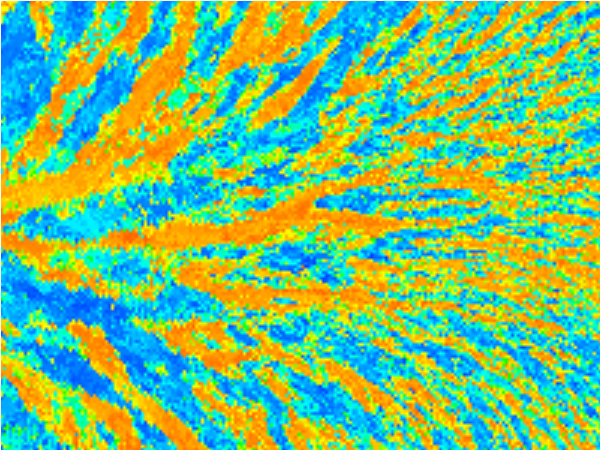}
& \qualimg{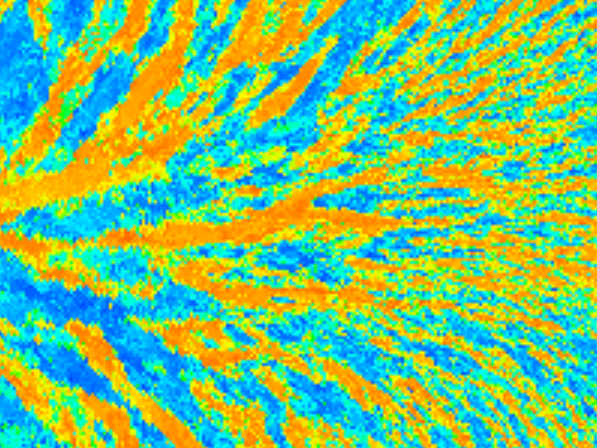}
& \qualimg{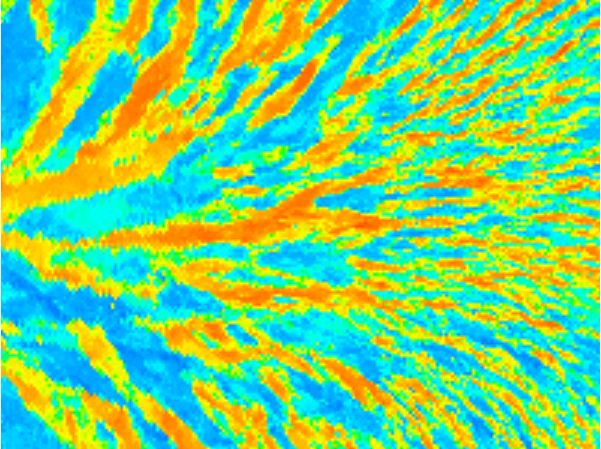}
& \qualimg{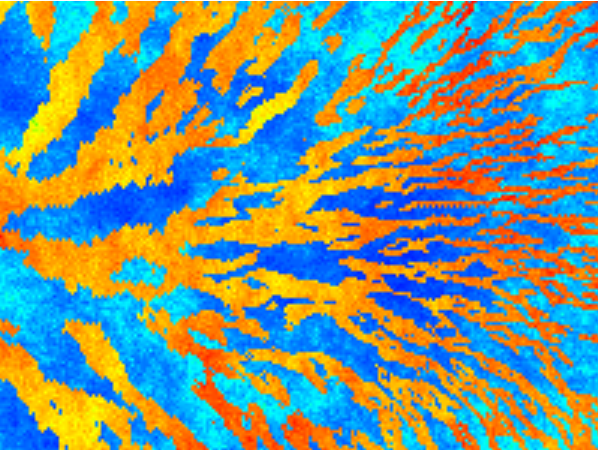}
\\[2pt]

\rowlab{Inline = 121}
& \qualimg{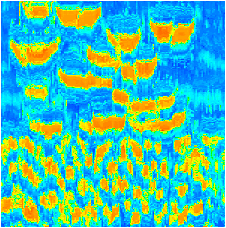}
& \qualimg{figures/picture/slice/KNN_InLine121.png}
& \qualimg{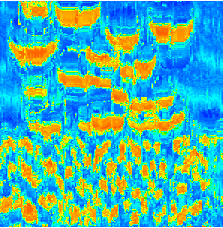}
& \qualimg{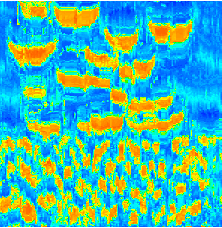}
& \qualimg{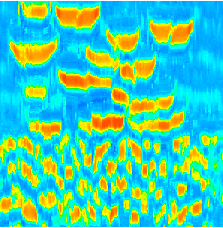}
& \qualimg{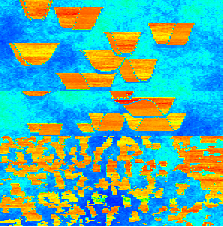}
\\[3pt]

& {\scriptsize XGBoost}
& {\scriptsize KNN}
& {\scriptsize Self-KNN}
& {\scriptsize COREG}
& {\scriptsize CoGameReg}
& {\scriptsize Ground Truth}
\end{tabular}

\caption{Qualitative comparison of P-wave velocity
predictions on representative slices of the synthetic
3D dataset. From top to bottom, the rows correspond
to Time = 9, Time = 110, Time = 188, and
Inline = 121. From left to right, the columns show
XGBoost, KNN, Self-KNN, COREG, CoGameReg, and the
ground truth. CoGameReg is placed next to the ground
truth for direct visual comparison. All panels use
the same P-wave velocity color scale as
Figure~\ref{fig:synthetic_inputs}(a).
}
\label{fig:qualitative_multi_slices}
\end{figure}

\subsection{Real Field Application in the Western Field Area}
\label{subsec:field_application}

\subsubsection{Geological setting and data}
\label{subsec:field_data}

The field data set was acquired from a proprietary tight-gas survey in
western China, hereafter referred to as the Western Field Area. To
comply with data-confidentiality requirements, the specific field name,
original well identifiers, and absolute survey coordinates are not
disclosed. The 18 wells used in this study are referred to as Well~1
through Well~18.

The target interval is a Permian tight-sandstone reservoir and
constitutes one of the principal exploration intervals in the study
area. The regional structure is generally gentle, although locally
developed low-relief structures and small-displacement faults are
present. The survey can be broadly divided into a western structural
low, a central slope, and an eastern uplift.

The target interval was deposited in a braided-river delta-plain
environment, with sediment supply predominantly from south to north.
The succession consists mainly of interbedded sandstone and mudstone.
The sandstone bodies are thin and laterally variable, generally less
than 15~m thick and averaging approximately 11~m. The gas-bearing
intervals are typically 2--8~m thick, and the distributary channels are
approximately 0.9--1.3~km wide. Moreover, sandstone development does
not necessarily correspond directly to gas-bearing intervals. These
characteristics make the spatial prediction of water saturation
particularly challenging. Water saturation is selected as the target property because it integrates
information from pore-fluid distribution and reservoir heterogeneity, while
its estimation from seismic observations remains challenging due to the
indirect relationship between seismic responses and fluid properties.

The available seismic data consist of a three-dimensional post-stack time-domain seismic volume, covering
approximately 140~km$^2$, with a bin size of 20~m $\times$ 20~m. The dominant frequency of the complete
seismic volume is approximately 30~Hz, whereas the dominant frequency
near the target interval ranges from approximately 35 to 45~Hz. The
processed seismic data exhibit a relatively high signal-to-noise ratio
and no obvious acquisition footprint within the target interval.

A total of 18 wells with valid water-saturation logs were retained for
the field experiment. The available well-log data include acoustic
transit-time, measured shear-wave, and ground truth water-saturation
logs. Before model construction, the well logs were subjected to
quality control, borehole-effect correction, outlier removal, and
interwell standardization. Synthetic seismograms were generated to
establish the well-to-seismic relationships before the logs were
resampled to the seismic sampling interval.

\subsubsection{Well-to-seismic calibration and experimental protocol}
\label{subsec:field_protocol}

The wells were partitioned at the well level, rather than at the sample
level, to evaluate spatial generalization between different well
locations. Well~1 through Well~10 were used for model training,
Well~11 through Well~15 were used for hyperparameter validation, and
Well~16 through Well~18 were reserved for blind-well evaluation. The
well partition is summarized in Table~\ref{tab:field_well_partition}.

\begin{table}[htbp]
\centering
\caption{Well partition used in the Western Field Area experiment.
The well names are anonymized for data-confidentiality reasons.}
\label{tab:field_well_partition}
\begin{tabular}{lcl}
\hline
Subset & Number & Anonymized well identifiers \\
\hline
Training   & 10 & Well~1--Well~10 \\
Validation & 5  & Well~11--Well~15 \\
Blind test & 3  & Well~16--Well~18 \\
\hline
\end{tabular}
\end{table}

Only the water-saturation samples from the training wells were used as
labeled data during model construction. The validation-well logs were
used only for hyperparameter selection and model configuration. The
blind-test-well logs were completely withheld from model training,
ensemble-weight estimation, pseudo-label generation, hyperparameter
selection, and model selection. They were accessed only after the
complete modeling workflow had been fixed.

\begin{figure}[p]
\centering

\newcommand{\fieldinputpanel}[2]{%
  \includegraphics[
    width=\linewidth,
    height=0.13\textheight,
    keepaspectratio
  ]{#1}
  \par\vspace{2pt}
  {\small #2\par}
  \vspace{6pt}
}

\fieldinputpanel{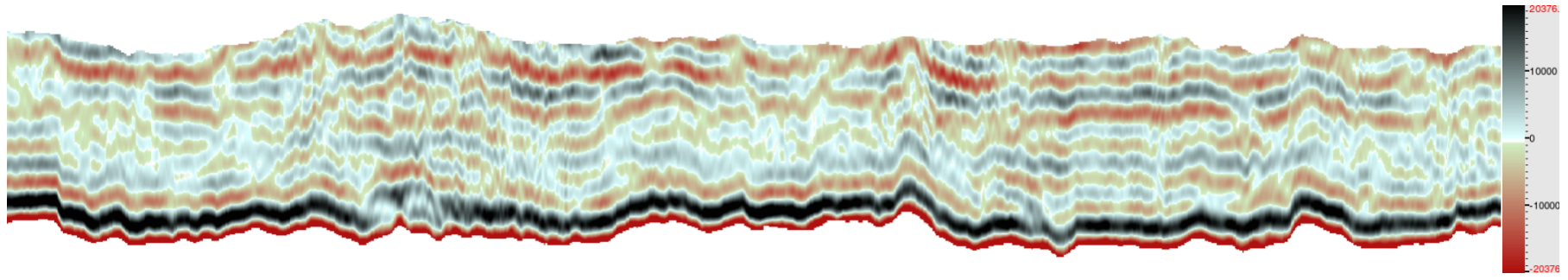}
{(a) Post-stack seismic}

\fieldinputpanel{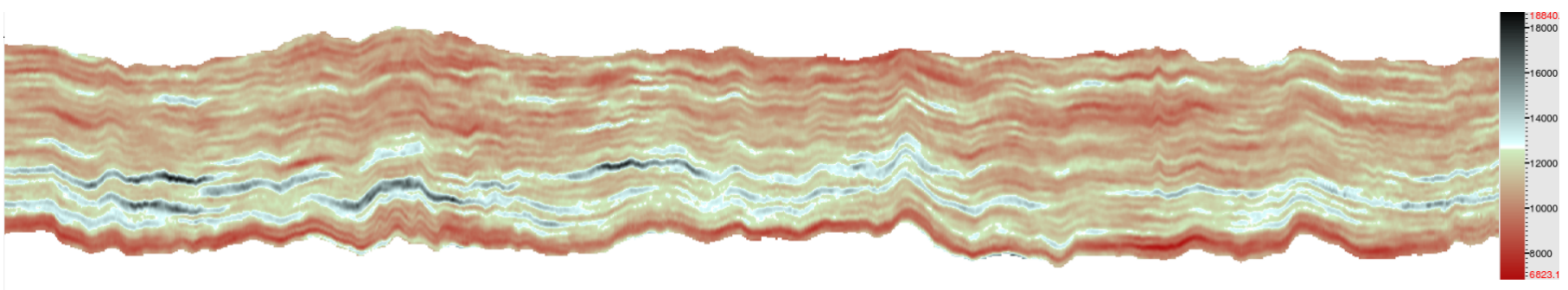}
{(b) P-wave impedance: post-stack geostatistical inversion}

\fieldinputpanel{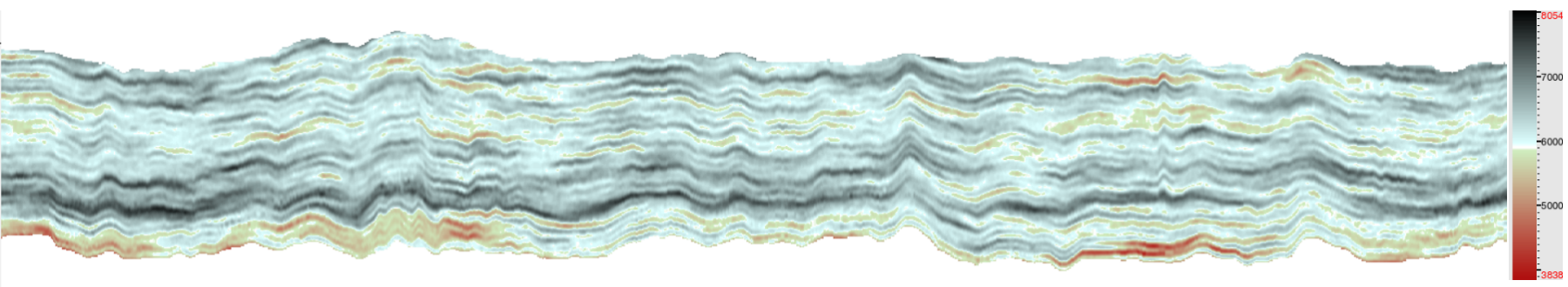}
{(c) S-wave impedance: pre-stack geostatistical inversion}

\fieldinputpanel{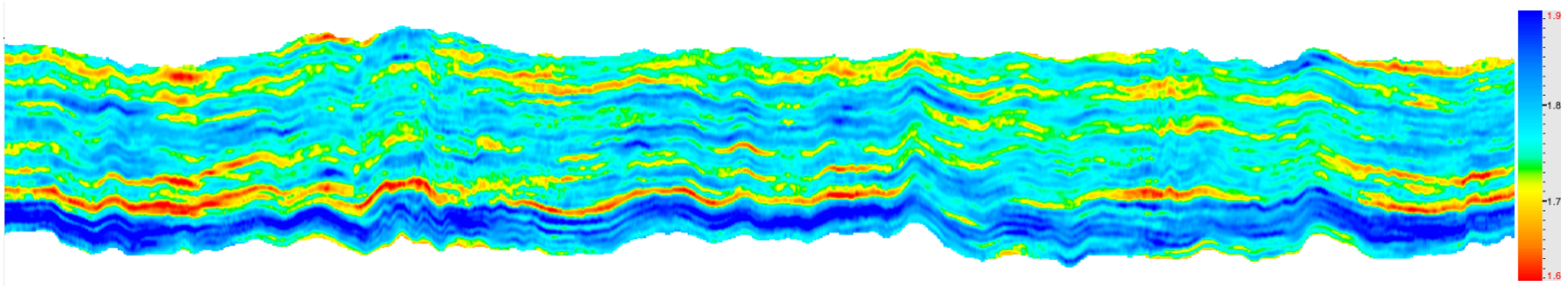}
{(d) $V_p/V_s$: pre-stack geostatistical inversion}

\fieldinputpanel{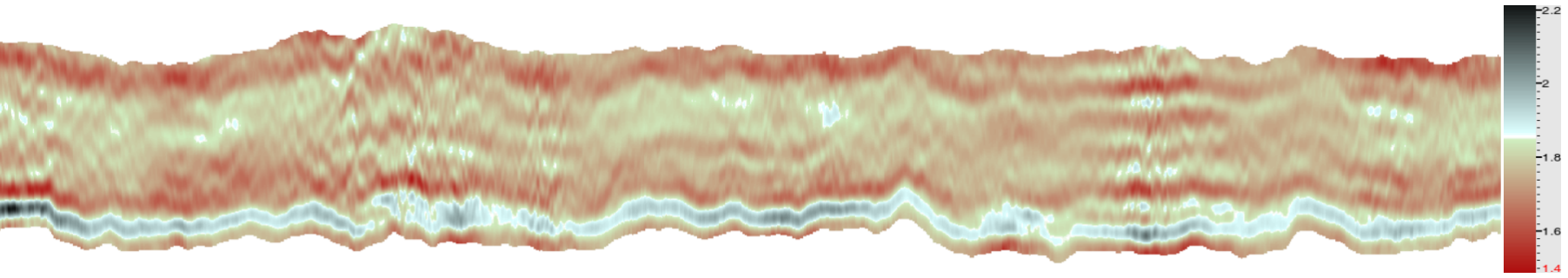}
{(e) $V_p/V_s$: pre-stack Bayesian inversion}

\caption{Vertical sections of the five seismic-derived
volumes used as input features for water-saturation
prediction in the Western Field Area. Each panel is
displayed with its own color scale.}
\label{fig:field_input_sections}
\end{figure}

Five seismic-derived volumes were used as input features
for water-saturation prediction: the post-stack seismic
volume, P-wave impedance from post-stack geostatistical
inversion, S-wave impedance from pre-stack geostatistical
inversion, and two $V_p/V_s$ volumes obtained from
pre-stack geostatistical and Bayesian inversion,
respectively. The post-stack seismic volume represents
the reflection response, whereas the inversion-derived
volumes provide estimates of impedance and velocity-ratio
properties. The two $V_p/V_s$ inputs describe the same
elastic-property ratio through different inversion
procedures. Their joint use allows the regression
framework to combine these seismic-derived descriptions
when estimating water saturation. The input volumes were obtained through seismic processing and
inversion using GeoEast. These precomputed volumes were used as the common inputs
to XGBoost, COREG, UCVME, and CoGameReg. The field experiment
therefore evaluates the regression frameworks for water-saturation
prediction given the same seismic-derived information. Sections of the five input volumes along a common arbitrary
seismic line are shown in Figure~\ref{fig:field_input_sections}.
The same line is used for the XGBoost, COREG, UCVME, and
CoGameReg prediction sections in
Figure~\ref{fig:field_vertical_section}, and its trajectory
is marked by the red polyline in
Figure~\ref{fig:field_horizon_slice}. This common profile
provides a direct spatial reference for examining the input
attributes alongside the predicted water-saturation distributions.

Seismic samples without corresponding water-saturation labels were treated
as unlabeled data. These samples formed $D_U$, from which the bounded
locality-aware candidate pool $U'_t$ was initialized and progressively
refilled during collaborative learning. The trained model was subsequently
applied to the complete seismic volume. After preprocessing, the field
experiment contained 1,248 labeled training samples, 625 validation samples,
393 blind-test samples, and approximately $8.2\times10^7$ unlabeled seismic
samples, illustrating the pronounced imbalance between the available
well-log labels and the large-scale seismic observations. 
For the field experiment, the base-model pool comprised XGBoost, SVR,
MLP, Ridge regression, and KNN regression. The complete CoGameReg
configuration is provided in Appendix~B.2 of the supplementary material.

The field comparison includes XGBoost as a supervised
baseline, COREG as a classical co-training regression
baseline, and UCVME as a modern semi-supervised baseline.
UCVME is included because it achieved the lowest RMSE and
highest $R^2$ among the competing methods in the synthetic
experiment. All four methods use the same five input
volumes and training/validation/blind-test well partition.
XGBoost uses the labeled training samples for model
fitting, whereas the semi-supervised methods additionally
use unlabeled seismic samples. Their predictions are
evaluated using the same valid-sample and output
post-processing rules. The COREG parameter settings used
in the synthetic and field experiments are provided in
Appendix~B.2. The spatial predictions are compared on the
same vertical section and reservoir horizon in
Figures~\ref{fig:field_vertical_section}
and~\ref{fig:field_horizon_slice}, respectively, to examine
differences in the spatial distribution and continuity
of the predicted water saturation.

Because CoGameReg is designed as a general semi-supervised regression
framework rather than a water-saturation-specific model, no physical range
constraint was imposed during model training. In the field application, the
predicted values were post-processed according to the physical definition of
water saturation. Specifically, predictions from all four methods were
clipped to the valid range of 0--100\% before visualization
and quantitative evaluation on the validation and
blind-test wells.
This post-processing step only enforces the valid physical range of the target property and does not affect the learning procedure or pseudo-label generation strategy of CoGameReg.

\begin{figure}[htbp]
\centering
\includegraphics[width=0.6\columnwidth]{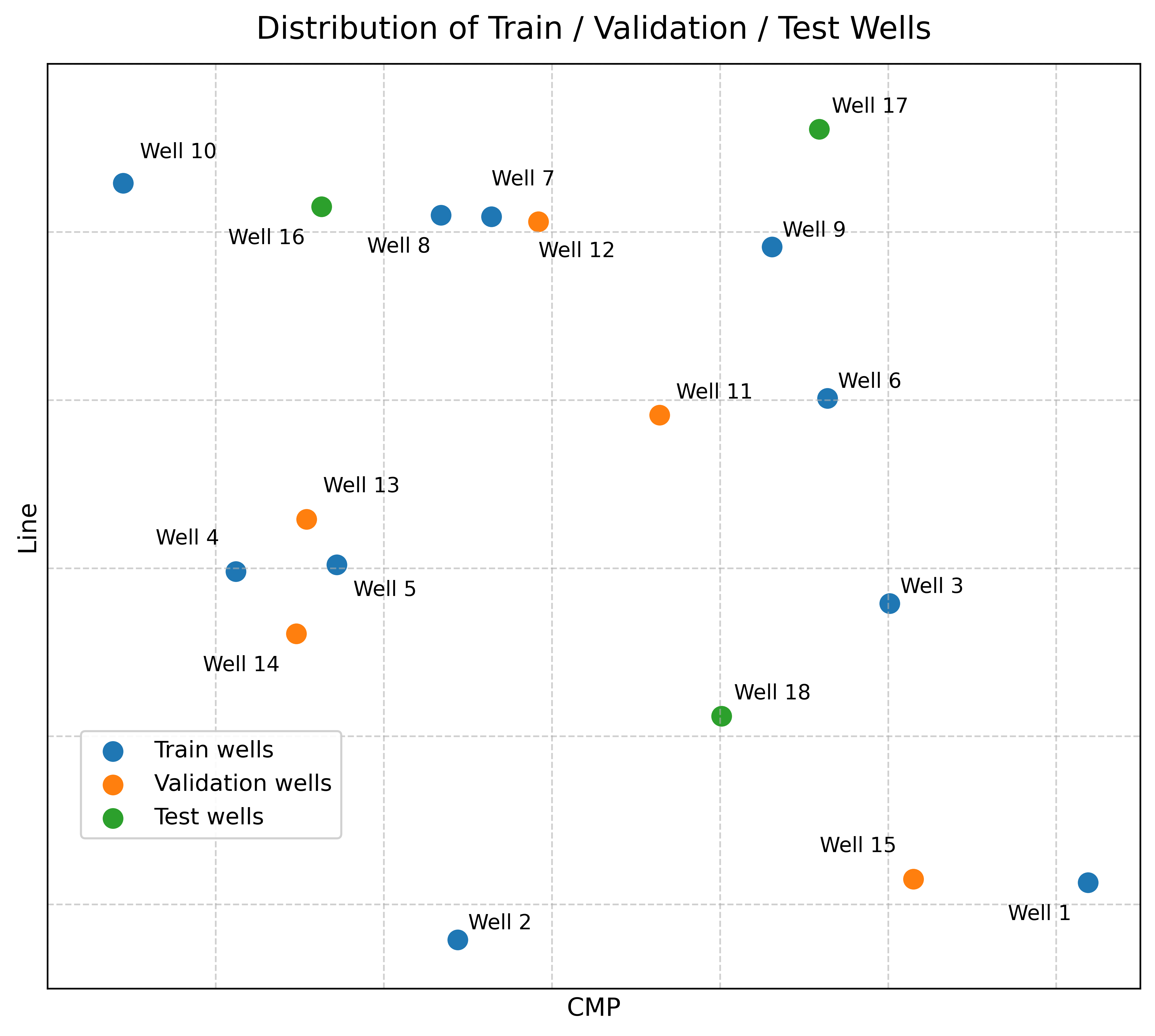}
\caption{Spatial distribution and data partition of the 18 anonymized
wells in the Western Field Area. Well~1--Well~10 were used for
training, Well~11--Well~15 for validation, and Well~16--Well~18 for
blind-well evaluation. Different colors denote the three
well subsets.}
\label{fig:field_well_locations}
\end{figure}

Figure~\ref{fig:field_well_locations} shows the anonymized spatial
distribution of the 18 wells. The training, validation, and blind-test
wells are distributed across different parts of the survey, allowing
the field experiment to evaluate the spatial generalization of
CoGameReg between different well locations.

\subsubsection{Three-dimensional prediction}
\label{subsec:field_3d_prediction}

The trained CoGameReg model was applied to the seismic
volume to generate spatially distributed water-saturation
estimates. Predictions from XGBoost, COREG, and UCVME provide
references for examining the resulting spatial patterns. The vertical
and horizon-based comparisons are presented below,
whereas blind-well predictive accuracy is evaluated
separately in Section~\ref{subsec:field_blind_evaluation}.

\begin{figure}[htbp]
\centering

\includegraphics[width=\linewidth]
{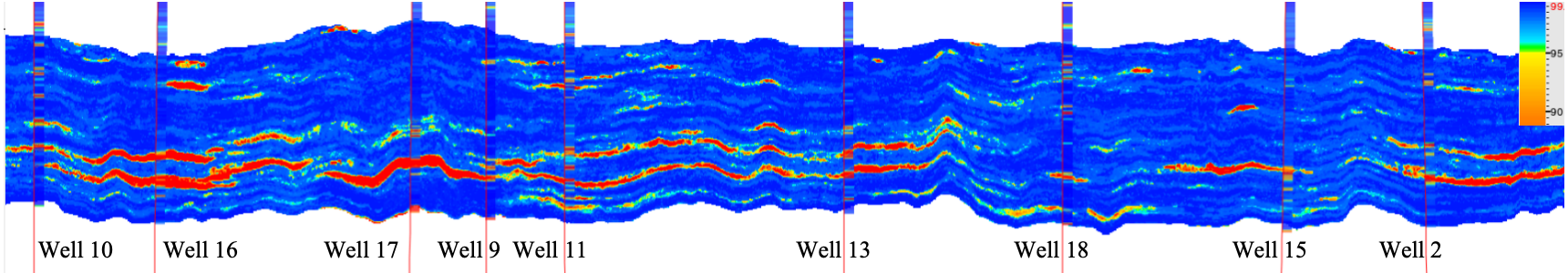}
\par\vspace{3pt}
{\small (a) XGBoost\par}

\vspace{10pt}

\includegraphics[width=\linewidth]
{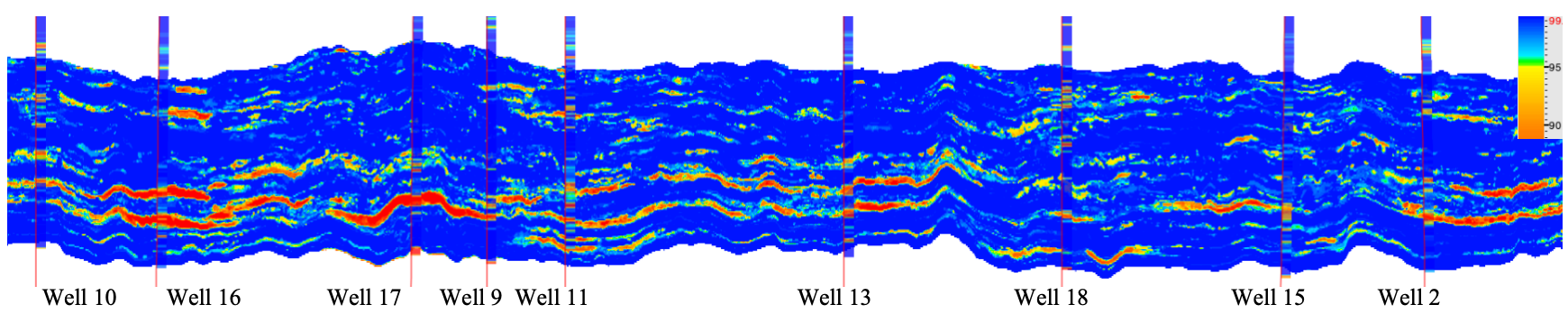}
\par\vspace{3pt}
{\small (b) COREG\par}

\vspace{10pt}

\includegraphics[width=\linewidth]
{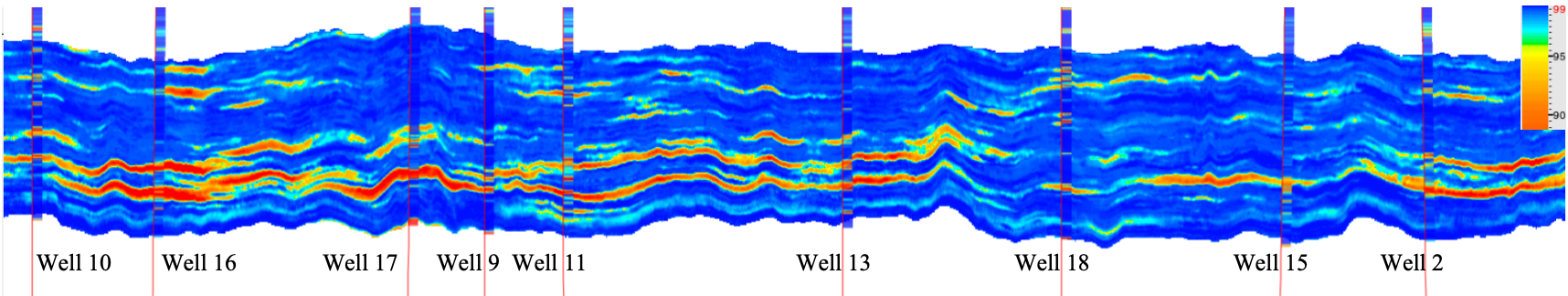}
\par\vspace{3pt}
{\small (c) UCVME\par}

\vspace{10pt}

\includegraphics[width=\linewidth]
{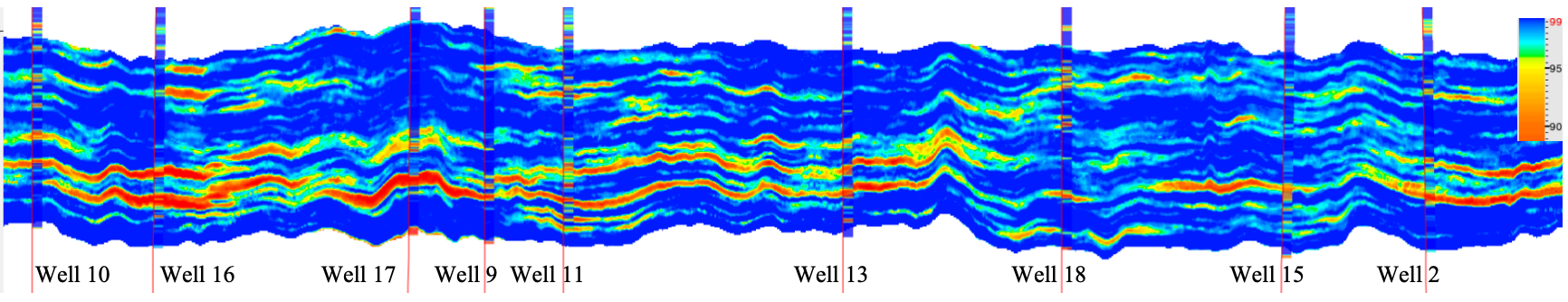}
\par\vspace{3pt}
{\small (d) CoGameReg\par}

\caption{Comparison of time-domain water-saturation
sections predicted by (a) XGBoost, (b) COREG,
(c) UCVME, and (d) CoGameReg along the same arbitrary
seismic line crossing nine wells in the Western Field
Area. Projected well locations and reference
water-saturation logs are displayed along the
sections. The line includes training, validation,
and blind-test wells.}
\label{fig:field_vertical_section}

\end{figure}

Figure~\ref{fig:field_vertical_section} compares the
predictions from XGBoost, COREG, UCVME, and CoGameReg
along an arbitrary seismic line crossing nine wells.
All four methods exhibit broadly layered saturation
patterns, with differences in the strength and continuity
of local anomalies. COREG contains more fragmented local
patches in parts of the displayed interval, whereas
UCVME and CoGameReg show more laterally continuous bands
in those areas. The projected
well locations and reference water-saturation tracks
provide spatial context for inspecting the predictions
near well control. Because the section includes training,
validation, and blind-test wells, the displayed well
tracks have different roles in model development and
evaluation.

\begin{figure}[htbp]
\centering

\begin{minipage}[t]{0.48\linewidth}
  \centering
  \vspace{0pt}
  \includegraphics[width=\linewidth]
  {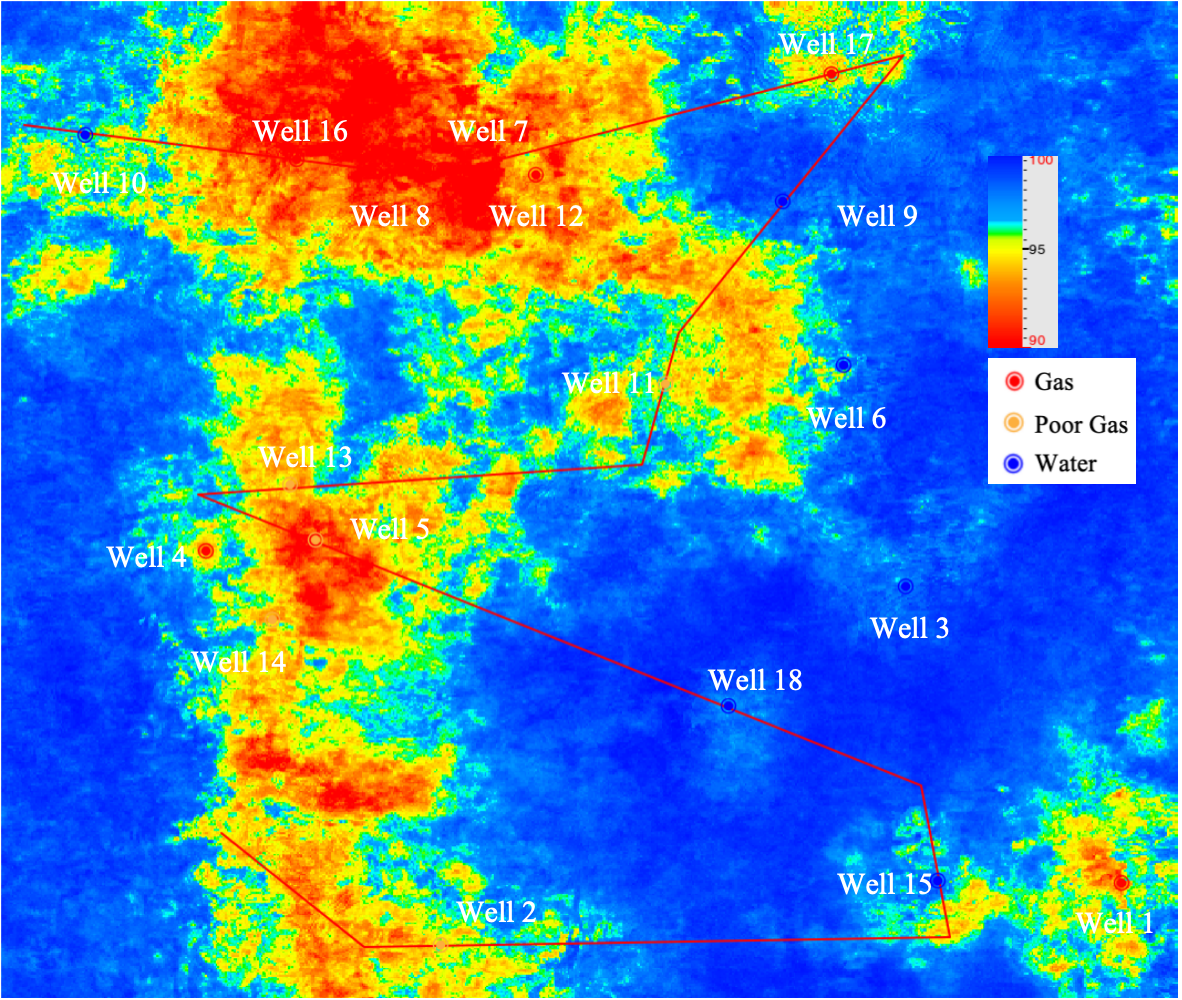}
  \par\vspace{4pt}
  {\small (a) XGBoost\par}
\end{minipage}
\hfill
\begin{minipage}[t]{0.48\linewidth}
  \centering
  \vspace{0pt}
  \includegraphics[width=\linewidth]
  {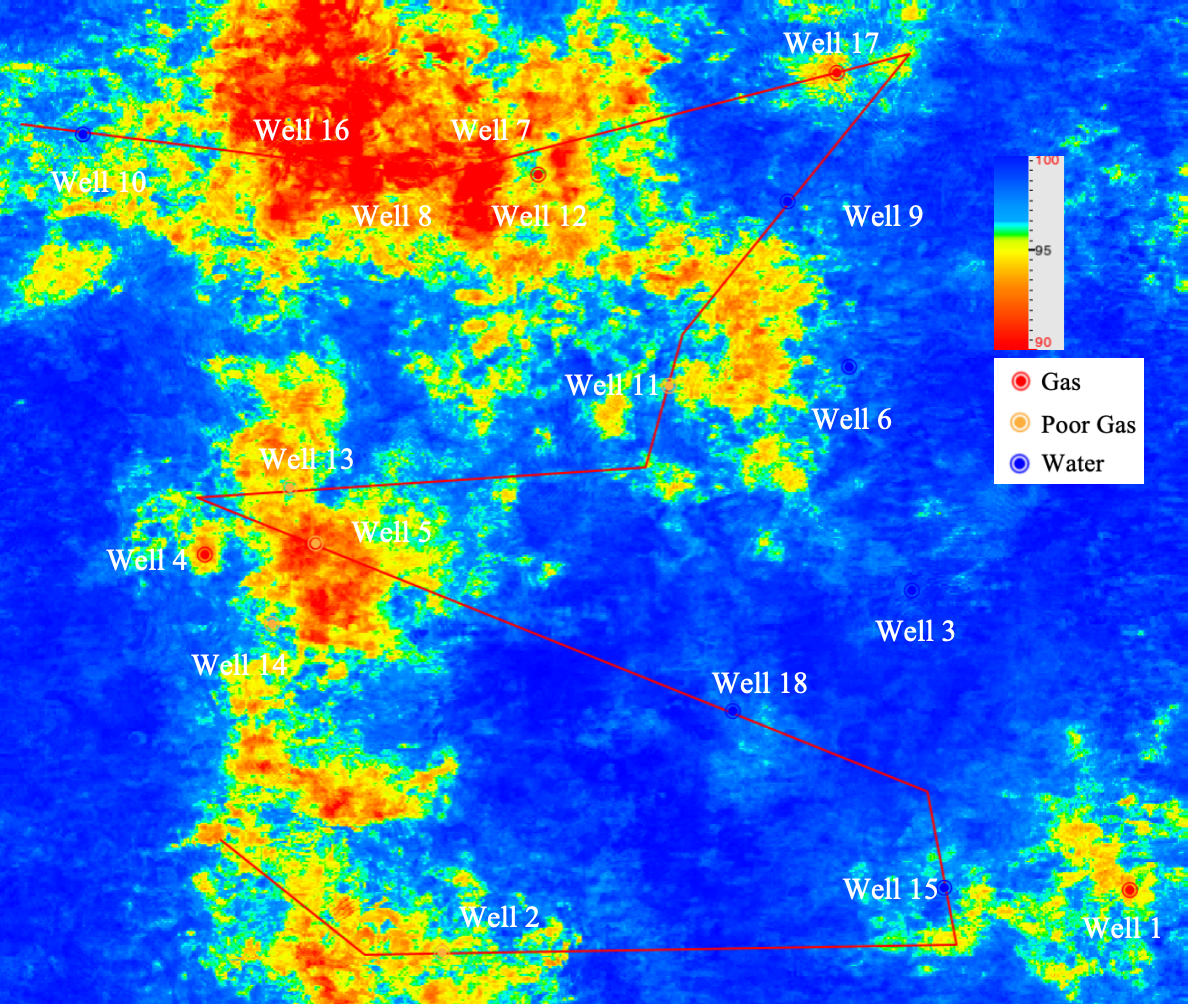}
  \par\vspace{4pt}
  {\small (b) COREG\par}
\end{minipage}

\par\vspace{12pt}

\begin{minipage}[t]{0.48\linewidth}
  \centering
  \vspace{0pt}
  \includegraphics[width=\linewidth]
  {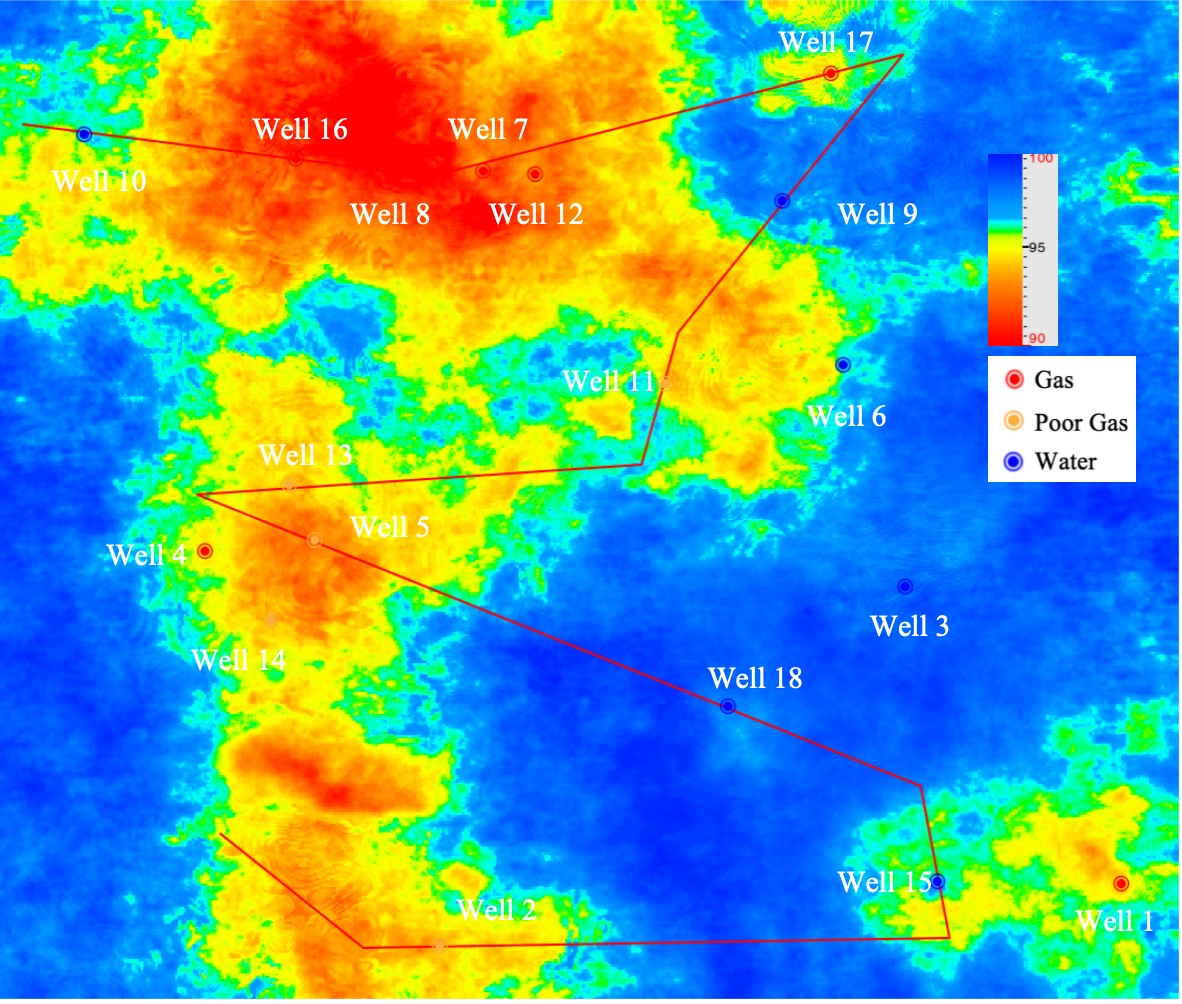}
  \par\vspace{4pt}
  {\small (c) UCVME\par}
\end{minipage}
\hfill
\begin{minipage}[t]{0.48\linewidth}
  \centering
  \vspace{0pt}
  \includegraphics[width=\linewidth]
  {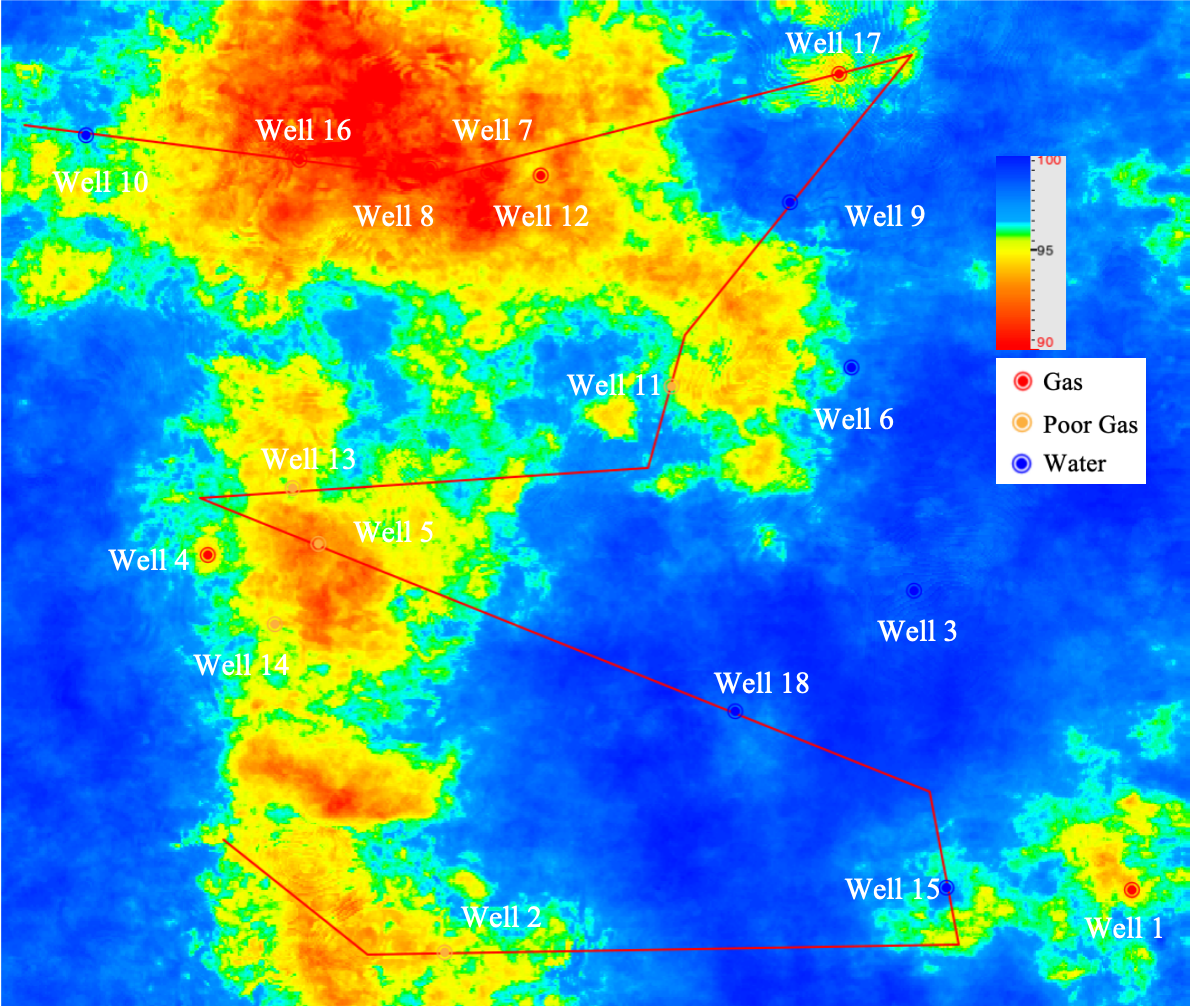}
  \par\vspace{4pt}
  {\small (d) CoGameReg\par}
\end{minipage}

\caption{Horizon-based water-saturation maps extracted
from the three-dimensional volumes predicted by
(a) XGBoost, (b) COREG, (c) UCVME, and (d) CoGameReg
in the Western Field Area. All four panels show the
same reservoir horizon and use the same spatial extent
and water-saturation color scale. The background colors
represent predicted water saturation. Colored well symbols
denote Gas, Poor Gas, and Water interpretations derived
from well logs at the corresponding horizon and projected
onto the maps for reference. The red polyline indicates
the trajectory of the vertical section shown in
Figure~\ref{fig:field_vertical_section}.}
\label{fig:field_horizon_slice}

\end{figure}

Figure~\ref{fig:field_horizon_slice} presents maps obtained
by sampling the three-dimensional water-saturation
volumes predicted by XGBoost, COREG, UCVME, and CoGameReg
along the same reservoir horizon.
The overlaid well symbols represent fluid interpretations
derived from well logs at the corresponding horizon,
categorized as Gas, Poor Gas, and Water. These interpretations
are projected onto the maps at the well locations to provide
geological reference information. For example, the gas
interpretation at Well~12 and the poor-gas interpretation at
Well~5 occur within relatively low predicted water-saturation
regions, whereas the water interpretations at Wells~3, 9,
and~18 occur within a higher-saturation background.

All four methods reproduce the broad spatial contrast
between relatively low and high saturation regions,
while their local spatial patterns differ. XGBoost and
COREG show more small-scale variations within and around
the main upper anomaly near Wells~16, 7, 8, and~12.
UCVME and CoGameReg show smoother and more connected
patterns in this region. Similar differences are visible
along the anomalous zone near Wells~13, 5, and~14. The well
interpretations provide geological context for these
predicted distributions, while the comparison between
methods highlights differences in their spatial continuity.
The red polyline marks the trajectory of the vertical
section shown in Figure~\ref{fig:field_vertical_section},
linking the horizon-based distributions to the vertical
prediction patterns.

These comparisons describe differences in the continuity
and morphology of the predicted saturation distributions.
They provide qualitative spatial context for the well-based
evaluation. Because reference water saturation is unavailable
away from the wells, greater visual continuity alone does
not establish higher prediction accuracy throughout the
seismic volume.

\subsubsection{Blind-well evaluation}
\label{subsec:field_blind_evaluation}

\begin{figure}[!htbp]
    \centering
    \includegraphics[width=0.5\linewidth]{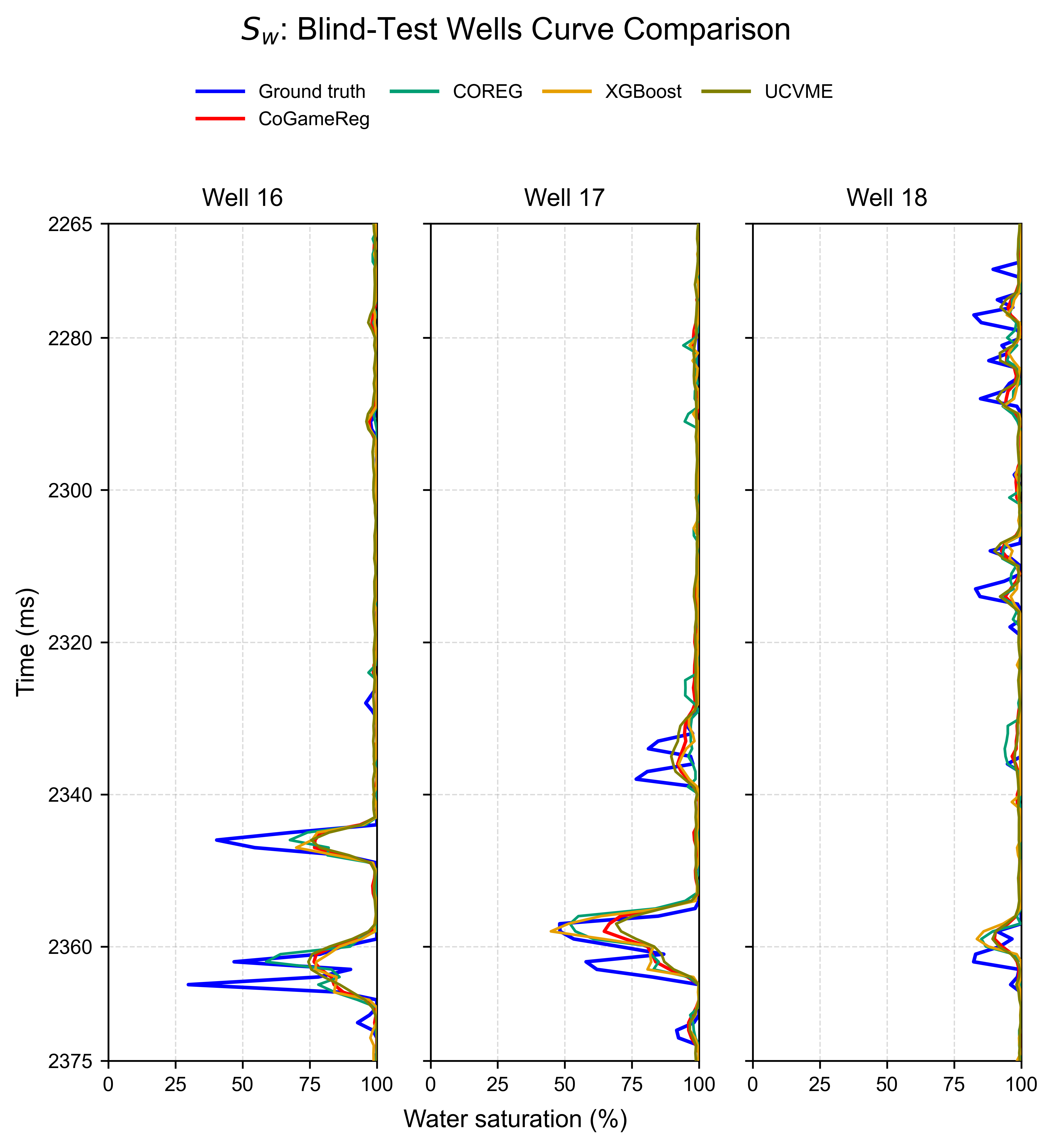}
    \caption{Comparison of ground-truth water-saturation curves
with predictions from XGBoost, COREG, UCVME, and CoGameReg
at the three blind-test wells (Wells~16--18) in the Western
Field Area.}
    \label{fig:field_blind_curves}
\end{figure}

The blind-well performance of XGBoost, COREG, UCVME, and
CoGameReg was evaluated at the same three blind-test wells,
which were excluded from model training and hyperparameter
selection. The predicted water-saturation curves were
sampled at the corresponding well locations and compared
with the ground truth water-saturation logs after
well-to-seismic calibration and temporal resampling.
Quantitative metrics were calculated over 2265--2409~ms,
inclusive, with 1~ms sampling, using samples valid for
the reference and all four methods. For visualization,
the curve comparisons in
Figure~\ref{fig:field_blind_curves} and Appendix~E show
the interval of 2265--2375~ms.

Following the physical post-processing described in subsection~\ref{subsec:field_protocol}, the
displayed predictions were bounded between 0\% and 100\% using the same
predefined clipping rule for all wells.

Figure~\ref{fig:field_blind_curves} compares the ground truth
water-saturation curves with predictions from XGBoost,
COREG, UCVME, and CoGameReg at the three blind wells. Overall,
CoGameReg reproduces the principal vertical variations and identifies
the major low-water-saturation intervals at previously unseen well
locations. The agreement nevertheless varies among wells, reflecting
the strong lateral heterogeneity of the thin tight-sandstone reservoir. Additional comparisons between the ground truth and predicted
water-saturation curves at the training and validation wells are
provided in Appendix~E of the supplementary material.

For CoGameReg, Well~16 and Well~17 have $R^2$ values of
0.5653 and 0.6748 and Pearson correlation coefficients
of 0.8326 and 0.8684, respectively, exceeding the
corresponding values at Well~18. These results indicate that the main vertical
variations of water saturation are reproduced at these two independent
well locations. At Well~18, the $R^2$ value is relatively
low at 0.2700, but its RMSE is only 3.6080 percentage points, which is
the lowest among the three blind wells. This combination is consistent
with the comparatively limited variation of the reference curve at
Well~18, because $R^2$ becomes sensitive to relatively small
prediction deviations when the target variance is low.
When all 393 valid blind-well samples are pooled, CoGameReg obtains
an $R^2$ of 0.5942, an RMSE of 5.7441 percentage points, an MAE of
2.1742 percentage points, and a Pearson correlation coefficient of
0.8265. The overall bias, defined as the mean difference between the
prediction and reference, is 0.7693 percentage points.

For CoGameReg, macro-average statistics were also calculated
by assigning equal weight to each of the three blind wells.
The macro-average $R^2$ is $0.5034 \pm 0.2094$, and the
macro-average RMSE is $5.5573 \pm 1.8027$ percentage points.
These values are reported as the mean $\pm$ sample standard
deviation across wells, describing the variation in
prediction performance among the three blind-well locations.

Table~\ref{tab:field_four_methods_pooled} compares the four
methods using common valid samples from the validation
and blind-test wells. On the validation wells, CoGameReg
achieves the lowest MAE (1.9746 percentage points) and
highest Pearson correlation (0.7807), while UCVME obtains
a slightly higher $R^2$ (0.5416 versus 0.5407) and lower
RMSE (4.4086 versus 4.4128 percentage points). On the
blind-test wells, CoGameReg has the highest Pearson
correlation (0.8265), whereas COREG achieves the highest
pooled $R^2$ (0.6148) and lowest RMSE (5.5969 percentage
points), and XGBoost has the lowest MAE (2.1533 percentage
points). These results show that no single method performs
best across all reported metrics on both subsets.

Taken together, the blind-well results demonstrate that CoGameReg
provides acceptable cross-well generalization for the real field data
and reproduces the major vertical water-saturation trends. However,
the variable performance among the blind wells indicates that the field
results should be interpreted as data-driven reservoir-property estimates.

\begin{table}[t]
\centering
\caption{Pooled water-saturation prediction results for four
representative methods in the field experiment.}
\label{tab:field_four_methods_pooled}

\small
\setlength{\tabcolsep}{5pt}
\renewcommand{\arraystretch}{1.12}

\begin{tabular}{lrrrrrr}
\toprule
Method & $N$ & $R^2\uparrow$ & RMSE$\downarrow$
       & MAE$\downarrow$ & Pearson $r\uparrow$ & Bias \\
\midrule

\multicolumn{7}{l}{\textit{Validation wells (Wells 11--15)}} \\

XGBoost
& 625 & 0.4456 & 4.8480 & 2.1656 & 0.6763 & 0.6556 \\

COREG
& 625 & 0.4121 & 4.9923 & 2.1673 & 0.6640 & 0.9781 \\

UCVME
& 625 & \textbf{0.5416} & \textbf{4.4086}
& 2.2239 & 0.7570 & 0.1965 \\

CoGameReg
& 625 & 0.5407 & 4.4128
& \textbf{1.9746} & \textbf{0.7807} & 0.6828 \\

\midrule
\multicolumn{7}{l}{\textit{Blind-test wells (Wells 16--18)}} \\

XGBoost
& 393 & 0.6052 & 5.6657
& \textbf{2.1533} & 0.7923 & 0.7121 \\

COREG
& 393 & \textbf{0.6148} & \textbf{5.5969}
& 2.1990 & 0.7908 & 0.5807 \\

UCVME
& 393 & 0.5486 & 6.0582 & 2.6311 & 0.7992 & 0.3638 \\

CoGameReg
& 393 & 0.5942 & 5.7441 & 2.1742
& \textbf{0.8265} & 0.7693 \\

\bottomrule
\end{tabular}

\par\smallskip
\begin{minipage}{\linewidth}
\footnotesize
\textit{Notes:}
$N$ denotes the number of common valid samples.
Bias is defined as
$\operatorname{mean}(\widehat{S}_w-S_w)$.
Pooled metrics are computed after concatenating the valid samples
across wells.
Bold indicates the best $R^2$, RMSE, MAE, or Pearson $r$ among the
four displayed methods within each subset; bias is not ranked.
\end{minipage}

\end{table}

\subsubsection{Geological interpretation and limitations}
\label{subsec:field_interpretation}

The field application addresses the prediction of water saturation in a thin, laterally variable tight-sandstone reservoir. The five input volumes provide seismic and inversion-derived information with different sensitivities to elastic properties. In particular, the two Vp/Vs volumes estimate the same physical quantity through different inversion procedures and may contain both complementary information and correlated errors. The Shapley-derived weights allocate influence among regressors; they do not by themselves remove attribute redundancy or resolve the nonuniqueness of estimating water saturation from seismic information. Cross-model agreement can therefore provide a useful fusion criterion while remaining susceptible to errors shared by the participating models.

The locality used for candidate selection is defined in seismic-attribute space. Similar attribute vectors can occur at spatially separated locations or in different lithologic and fluid settings. Consequently, a reduction in local proxy error does not by itself establish compatibility across geological facies. The fixed initial weights summarize regressor behavior at the labeled wells and do not provide location-dependent reliability estimates. Increasing distance from well control may coincide with changes in facies, rock properties, or seismic response, but the present blind-well comparison does not establish a quantitative relationship between prediction error and physical distance. The reported results should therefore be interpreted within the geological and attribute conditions represented by the available well control.

Finite seismic bandwidth also limits the interpretation of predicted saturation detail. The target interval contains sandstone bodies averaging approximately 11 m in thickness and gas-bearing intervals of 2–8 m, while the dominant frequency near the reservoir is approximately 35–45 Hz. Thin-bed interference and the difference in scale between well logs and seismic observations can contribute to smoothed boundaries and underestimated saturation contrasts. The 1 ms sampling used in the well comparison specifies the discretization interval, not the vertical resolving power of the seismic data. Fine-scale variations in the predicted saturation should therefore be interpreted with reference to the bandwidth, well-to-seismic calibration, and assumptions underlying the input volumes.

The maps and sections provide spatial context for the blind-well results. Because seismic and inversion volumes are themselves model inputs, agreement with their patterns is a consistency check rather than independent validation. Moreover, greater continuity can accompany smoothing of genuine geological contrasts. Reference water saturation is unavailable away from the well trajectories, so the present results do not quantify the accuracy of the complete prediction volume. CoGameReg provides a conditional reservoir-property estimate whose quantitative support comes from the blind-well comparison; its spatial interpretation should be checked against independent geological evidence.

\section{Conclusion}
This study developed CoGameReg for seismic reservoir property prediction under sparse well control. Its central contribution is the use of Shapley-derived model weights, estimated once from the labeled training data and then held fixed, in both shared pseudo-label construction and final prediction fusion. These weights determine how heterogeneous regressors contribute to the additional learning targets generated from unlabeled seismic attributes. Local error-reduction screening complements this model-level weighting by controlling candidate admission. This combination provides a practical way to use sparse well supervision together with dense seismic observations.
The public benchmark results show that CoGameReg achieves its largest performance advantage at the lowest labeled-data ratio, with the gap relative to competing methods narrowing as the proportion of labeled data increases. The benchmark ablations support placing contribution-aware weighting within the pseudo-labeling loop. CoGameReg achieves the best blind-well metrics in the synthetic experiment. In the field application, the clipped predictions reproduce major water-saturation trends at the three blind wells and yield the highest pooled Pearson correlation among the four compared methods. The field results therefore support applicability under the evaluated conditions, with performance depending on the metric and well location.

\section{Acknowledgements}
This work was supported by CNPC Science and Technology Major Project (No. SG0023Z0010050100111) and CNPC Youth Science and Technology Special Project "Research on Rock Physics and Facies-Driven Intelligent Seismic Reservoir Characterization Methods" (No. 2025DQ03007).

\section{Data and Materials Availability}
The source code for CoGameReg is available at
\url{https://github.com/ACE020215/CoGameReg}.
The public benchmark data sets and the Stanford VI-E synthetic reservoir data are available from the cited sources. The proprietary field seismic and well-log data are confidential and cannot be released.

\bibliographystyle{plain}
\bibliography{references_geophysics}

@book{ref_aki,
  author    = {Aki, K. and Richards, P. G.},
  title     = {Quantitative seismology},
  edition   = {2nd},
  publisher = {University Science Books},
  year      = {2002}
}

@book{ref_tarantola,
  author    = {Tarantola, A.},
  title     = {Inverse problem theory and methods for model parameter estimation},
  publisher = {Society for Industrial and Applied Mathematics},
  year      = {2005}
}

@article{ref_zoeppritz,
  author  = {Zoeppritz, K.},
  title   = {{VII b.} {\"U}ber Reflexion und Durchgang seismischer Wellen durch Unstetigkeitsfl{\"a}chen},
  journal = {Nachrichten von der Gesellschaft der Wissenschaften zu G\"ottingen, Mathematisch-Physikalische Klasse},
  volume  = {1919},
  pages   = {66--84},
  year    = {1919}
}

@article{ref_denoise,
  author  = {Saad, O. M. and Chen, Y.},
  title   = {Deep denoising autoencoder for seismic random noise attenuation},
  journal = {Geophysics},
  volume  = {85},
  number  = {4},
  pages   = {V367--V376},
  year    = {2020}
}

@article{ref_fault,
  author  = {Li, Y. and Wu, X. and Zhu, Z. and Ding, J. and Wang, Q.},
  title   = {{FaultSeg3D} plus: A comprehensive study on evaluating and improving {CNN}-based seismic fault segmentation},
  journal = {Geophysics},
  volume  = {89},
  number  = {5},
  pages   = {N77--N91},
  year    = {2024},
  doi     = {10.1190/geo2022-0778.1}
}

@article{ref_facies,
  author  = {Li, J. and Wu, X. and Ye, Y. and Yang, C. and Hu, Z. and Sun, X. and Zhao, T.},
  title   = {Unsupervised contrastive learning for seismic facies characterization},
  journal = {Geophysics},
  volume  = {88},
  number  = {1},
  pages   = {WA81--WA89},
  year    = {2023}
}

@article{ref_inversion,
  author  = {Song, L. and Yin, X. and Zhang, R. and Li, J. and Zhang, J. and Li, J.},
  title   = {Reservoir discrimination based on physic-informed semi-supervised learning},
  journal = {IEEE Transactions on Geoscience and Remote Sensing},
  volume  = {62},
  pages   = {4507314},
  year    = {2024},
  doi     = {10.1109/TGRS.2024.3409578}
}

@article{ref_multiseismic,
  author  = {Xu, M. and Zhao, L. and Gao, S. and Zhu, X. and Geng, J.},
  title   = {Joint use of multiseismic information for lithofacies prediction via supervised convolutional neural networks},
  journal = {Geophysics},
  volume  = {87},
  number  = {5},
  pages   = {M151--M162},
  year    = {2022}
}

@article{ref_gas_prediction_dl,
  author  = {Yang, J. and Lin, N. and Zhang, K. and Fu, C. and Zhang, C.},
  title   = {Transfer learning-based hybrid deep learning method for gas-bearing distribution prediction with insufficient training samples and uncertainty analysis},
  journal = {Energy},
  volume  = {299},
  pages   = {131414},
  year    = {2024},
  doi     = {10.1016/j.energy.2024.131414}
}

@article{ref_selfsupervised_gas,
  author  = {Han, L. and Wu, X. and Chen, R. and Shi, Y. and Hu, Z. and Fang, H.},
  title   = {Self-supervised pre-training and few-shot finetuning for gas-bearing prediction},
  journal = {Journal of Geophysical Research: Machine Learning and Computation},
  volume  = {2},
  pages   = {e2025JH000631},
  year    = {2025},
  doi     = {10.1029/2025JH000631}
}

@book{ref1,
  editor    = {Chapelle, O. and Sch\"olkopf, B. and Zien, A.},
  title     = {Semi-supervised learning},
  publisher = {The MIT Press},
  year      = {2006}
}

@inproceedings{ref2,
  author    = {Goldman, S. A. and Zhou, Y.},
  title     = {Enhancing supervised learning with unlabeled data},
  booktitle = {Proceedings of the 17th International Conference on Machine Learning},
  pages     = {327--334},
  year      = {2000}
}

@article{ref11,
  author  = {Kostopoulos, G. and Karlos, S. and Kotsiantis, S. and Ragos, O.},
  title   = {Semi-supervised regression: A recent review},
  journal = {Journal of Intelligent \& Fuzzy Systems},
  volume  = {35},
  pages   = {1483--1500},
  year    = {2018}
}

@inproceedings{ref5,
  author    = {Wei, C. and Sohn, K. and Mellina, C. and Yuille, A. and Yang, F.},
  title     = {{CReST}: A class-rebalancing self-training framework for imbalanced semi-supervised learning},
  booktitle = {Proceedings of the IEEE/CVF Conference on Computer Vision and Pattern Recognition},
  pages     = {10857--10866},
  year      = {2021}
}

@article{ref6,
  author  = {Li, J. and Zhu, Q.},
  title   = {A boosting self-training framework based on instance generation with natural neighbors for k nearest neighbor},
  journal = {Applied Intelligence},
  volume  = {50},
  pages   = {3535--3553},
  year    = {2020}
}

@article{ref7,
  author  = {Miyato, T. and Maeda, S.-I. and Koyama, M. and Ishii, S.},
  title   = {Virtual adversarial training: A regularization method for supervised and semi-supervised learning},
  journal = {IEEE Transactions on Pattern Analysis and Machine Intelligence},
  volume  = {41},
  pages   = {1979--1993},
  year    = {2019}
}

@article{ref8,
  author  = {Verma, V. and Kawaguchi, K. and Lamb, A. and Kannala, J. and Solin, A. and Bengio, Y. and Lopez-Paz, D.},
  title   = {Interpolation consistency training for semi-supervised learning},
  journal = {Neural Networks},
  volume  = {145},
  pages   = {90--106},
  year    = {2022}
}

@inproceedings{ref_graph_ssl,
  author    = {Zhu, X. and Ghahramani, Z. and Lafferty, J. D.},
  title     = {Semi-supervised learning using {G}aussian fields and harmonic functions},
  booktitle = {Proceedings of the 20th International Conference on Machine Learning},
  pages     = {912--919},
  publisher = {AAAI Press},
  year      = {2003}
}

@article{ref13,
  author  = {Dong, H. and Yang, L. and Wang, X.},
  title   = {Robust semi-supervised support vector machines with {L}aplace kernel-induced correntropy loss functions},
  journal = {Applied Intelligence},
  volume  = {51},
  pages   = {819--833},
  year    = {2021},
  doi     = {10.1007/s10489-020-01865-3}
}

@inproceedings{ref3,
  author    = {Blum, A. and Mitchell, T. M.},
  title     = {Combining labeled and unlabeled data with co-training},
  booktitle = {Proceedings of the 11th Annual Conference on Computational Learning Theory},
  pages     = {92--100},
  publisher = {Association for Computing Machinery},
  year      = {1998},
  doi       = {10.1145/279943.279962}
}

@article{ref35,
  author  = {Zhou, Z.-H. and Li, M.},
  title   = {Semisupervised regression with cotraining-style algorithms},
  journal = {IEEE Transactions on Knowledge and Data Engineering},
  volume  = {19},
  pages   = {1479--1493},
  year    = {2007},
  doi     = {10.1109/TKDE.2007.190644}
}

@article{ref14,
  author  = {Balcan, M.-F. and Blum, A.},
  title   = {A discriminative model for semi-supervised learning},
  journal = {Journal of the ACM},
  volume  = {57},
  pages   = {19:1--19:46},
  year    = {2010},
  doi     = {10.1145/1706591.1706599}
}

@article{ref15,
  author  = {Yang, T. and Priebe, C. E.},
  title   = {The effect of model misspecification on semi-supervised classification},
  journal = {IEEE Transactions on Pattern Analysis and Machine Intelligence},
  volume  = {33},
  pages   = {2093--2103},
  year    = {2011},
  doi     = {10.1109/TPAMI.2011.45}
}

@article{ref16,
  author  = {Li, Y.-F. and Tsang, I. W. and Kwok, J. T. and Zhou, Z.-H.},
  title   = {Convex and scalable weakly labeled {SVM}s},
  journal = {Journal of Machine Learning Research},
  volume  = {14},
  pages   = {2151--2188},
  year    = {2013}
}

@inproceedings{ref17,
  author    = {Li, Y.-F. and Zha, H.-W. and Zhou, Z.-H.},
  title     = {Learning safe prediction for semi-supervised regression},
  booktitle = {Proceedings of the 31st AAAI Conference on Artificial Intelligence},
  pages     = {2217--2223},
  year      = {2017}
}

@inproceedings{ref18,
  author    = {Ma, F. and Meng, D. and Xie, Q. and Li, Z. and Dong, X.},
  title     = {Self-paced co-training},
  booktitle = {Proceedings of the 34th International Conference on Machine Learning},
  series    = {Proceedings of Machine Learning Research},
  volume    = {70},
  pages     = {2275--2284},
  year      = {2017}
}

@inproceedings{ref39,
  author    = {Min, F. and Li, Y. and Liu, L.},
  title     = {Self-paced safe co-training for regression},
  booktitle = {Advances in Knowledge Discovery and Data Mining: Proceedings of the 26th Pacific-Asia Conference on Knowledge Discovery and Data Mining},
  series    = {Lecture Notes in Computer Science},
  volume    = {13281},
  pages     = {71--82},
  publisher = {Springer},
  year      = {2022},
  doi       = {10.1007/978-3-031-05936-0_6}
}

@article{ref_S2RMS,
  author  = {Liu, L. and Zhang, J. and Qian, K. and Min, F.},
  title   = {Semi-supervised regression via embedding space mapping and pseudo-label smearing},
  journal = {Applied Intelligence},
  volume  = {54},
  pages   = {9622--9640},
  year    = {2024},
  doi     = {10.1007/s10489-024-05686-6}
}

@inproceedings{ref37,
  author    = {Abdel Hady, M. F. and Schwenker, F. and Palm, G.},
  title     = {Semi-supervised learning for regression with co-training by committee},
  booktitle = {Artificial Neural Networks---ICANN 2009: Proceedings of the 19th International Conference, Part I},
  series    = {Lecture Notes in Computer Science},
  volume    = {5768},
  pages     = {121--130},
  publisher = {Springer},
  year      = {2009},
  doi       = {10.1007/978-3-642-04274-4_13}
}

@article{ref40,
  author  = {Lu, S. and Shi, X. and Li, M. and Jiao, J. and Feng, L. and Wang, G.},
  title   = {Semi-supervised random forest regression model based on co-training and grouping with information entropy for evaluation of depression symptoms severity},
  journal = {Mathematical Biosciences and Engineering},
  volume  = {18},
  pages   = {4586--4602},
  year    = {2021},
  doi     = {10.3934/mbe.2021233}
}

@article{ref31,
  author  = {Zhou, Z.-H. and Li, M.},
  title   = {Tri-training: Exploiting unlabeled data using three classifiers},
  journal = {IEEE Transactions on Knowledge and Data Engineering},
  volume  = {17},
  pages   = {1529--1541},
  year    = {2005},
  doi     = {10.1109/TKDE.2005.186}
}

@article{ref32,
  author  = {Gong, Y. and Wu, Q.},
  title   = {{SIVLC}: Improving the performance of co-training by sufficient-irrelevant views and label consistency},
  journal = {Applied Intelligence},
  volume  = {53},
  pages   = {20710--20729},
  year    = {2023},
  doi     = {10.1007/s10489-023-04611-7}
}

@incollection{ref33,
  author    = {Qiao, S. and Shen, W. and Zhang, Z. and Wang, B. and Yuille, A.},
  title     = {Deep co-training for semi-supervised image recognition},
  booktitle = {Computer Vision---ECCV 2018: Proceedings of the 15th European Conference},
  editor    = {Ferrari, V. and Hebert, M. and Sminchisescu, C. and Weiss, Y.},
  series    = {Lecture Notes in Computer Science},
  volume    = {11219},
  pages     = {142--159},
  publisher = {Springer},
  year      = {2018},
  doi       = {10.1007/978-3-030-01267-0_9}
}

@inproceedings{ref34,
  author    = {Chen, D.-D. and Wang, W. and Gao, W. and Zhou, Z.-H.},
  title     = {Tri-net for semi-supervised deep learning},
  booktitle = {Proceedings of the Twenty-Seventh International Joint Conference on Artificial Intelligence},
  pages     = {2014--2020},
  year      = {2018},
  doi       = {10.24963/ijcai.2018/278}
}

@inproceedings{ref42,
  author    = {Shen, Z. and Cao, P. and Yang, H. and Liu, X. and Yang, J. and Zaiane, O. R.},
  title     = {Co-training with high-confidence pseudo labels for semi-supervised medical image segmentation},
  booktitle = {Proceedings of the Thirty-Second International Joint Conference on Artificial Intelligence},
  pages     = {4199--4207},
  year      = {2023},
  doi       = {10.24963/ijcai.2023/467}
}

@article{ref44,
  author  = {Li, Q. and Chen, Y. and He, X. and Huang, L.},
  title   = {Co-training transformer for remote sensing image classification, segmentation, and detection},
  journal = {IEEE Transactions on Geoscience and Remote Sensing},
  volume  = {62},
  pages   = {5606218},
  year    = {2024},
  doi     = {10.1109/TGRS.2024.3354783}
}

@inproceedings{ref46,
  author    = {Liu, Z. and Ma, Q. and Ma, P. and Wang, L.},
  title     = {Temporal-frequency co-training for time series semi-supervised learning},
  booktitle = {Proceedings of the AAAI Conference on Artificial Intelligence},
  volume    = {37},
  pages     = {8923--8931},
  year      = {2023},
  doi       = {10.1609/aaai.v37i7.26072}
}

@article{ref49,
  author  = {Li, Y. and Xiong, H. and Wang, Q. and Kong, L. and Liu, H. and Li, H. and Bian, J. and Wang, S. and Chen, G. and Dou, D. and Yin, D.},
  title   = {{COLTR}: Semi-supervised learning to rank with co-training and over-parameterization for web search},
  journal = {IEEE Transactions on Knowledge and Data Engineering},
  volume  = {35},
  pages   = {12542--12555},
  year    = {2023},
  doi     = {10.1109/TKDE.2023.3270750}
}

@incollection{shapley1953value,
  author    = {Shapley, L. S.},
  title     = {A value for n-person games},
  booktitle = {Contributions to the theory of games, volume II},
  editor    = {Kuhn, H. W. and Tucker, A. W.},
  series    = {Annals of Mathematics Studies},
  volume    = {28},
  pages     = {307--317},
  publisher = {Princeton University Press},
  year      = {1953}
}

@inproceedings{ref_shap,
  author    = {Lundberg, S. M. and Lee, S.-I.},
  title     = {A unified approach to interpreting model predictions},
  booktitle = {Advances in Neural Information Processing Systems},
  volume    = {30},
  pages     = {4765--4774},
  year      = {2017}
}

@inproceedings{ghorbani2019data,
  author    = {Ghorbani, A. and Zou, J.},
  title     = {Data {Shapley}: Equitable valuation of data for machine learning},
  booktitle = {Proceedings of the 36th International Conference on Machine Learning},
  series    = {Proceedings of Machine Learning Research},
  volume    = {97},
  pages     = {2242--2251},
  year      = {2019}
}

@article{ref_BiGC,
  author  = {Garg, V. K. and Narahari, Y. and Narasimha Murty, M.},
  title   = {Novel biobjective clustering ({BiGC}) based on cooperative game theory},
  journal = {IEEE Transactions on Knowledge and Data Engineering},
  volume  = {25},
  pages   = {1070--1082},
  year    = {2013},
  doi     = {10.1109/TKDE.2012.73}
}

@misc{uci_repo,
  author       = {Kelly, M. and Longjohn, R. and Nottingham, K.},
  title        = {The {UCI} Machine Learning Repository},
  year         = {2023},
  note         = {University of California, Irvine, \url{https://archive.ics.uci.edu/}}
}

@misc{delve_repo,
  author       = {{DELVE Development Group}},
  title        = {{DELVE}: Data for Evaluating Learning in Valid Experiments},
  year         = {n.d.},
  note         = {University of Toronto, \url{https://www.cs.toronto.edu/~delve/}}
}

@misc{statlib_repo,
  author       = {{Carnegie Mellon University}},
  title        = {{StatLib} Datasets Archive},
  year         = {n.d.},
  note         = {\url{https://lib.stat.cmu.edu/datasets/}}
}

@inproceedings{gorishniy2025tabm,
  author    = {Gorishniy, Y. and Kotelnikov, A. and Babenko, A.},
  title     = {{TabM}: Advancing Tabular Deep Learning with Parameter-Efficient Ensembling},
  booktitle = {International Conference on Learning Representations},
  year      = {2025},
  url       = {https://proceedings.iclr.cc/paper_files/paper/2025/hash/c1ba41c694834aeef91ae161711d4939-Abstract-Conference.html}
}

@inproceedings{tarvainen2017meanteacher,
  author    = {Tarvainen, A. and Valpola, H.},
  title     = {Mean teachers are better role models: Weight-averaged consistency targets improve semi-supervised deep learning results},
  booktitle = {Advances in Neural Information Processing Systems},
  volume    = {30},
  pages     = {1195--1204},
  year      = {2017}
}

@inproceedings{jean2018ssdk,
  author    = {Jean, N. and Xie, S. M. and Ermon, S.},
  title     = {Semi-supervised deep kernel learning: Regression with unlabeled data by minimizing predictive variance},
  booktitle = {Advances in Neural Information Processing Systems},
  volume    = {31},
  pages     = {5322--5333},
  year      = {2018}
}

@inproceedings{dai2023ucvme,
  author    = {Dai, W. and Li, X. and Cheng, K.-T.},
  title     = {Semi-supervised deep regression with uncertainty consistency and variational model ensembling via {B}ayesian neural networks},
  booktitle = {Proceedings of the AAAI Conference on Artificial Intelligence},
  volume    = {37},
  pages     = {7304--7313},
  year      = {2023},
  doi       = {10.1609/aaai.v37i6.25890}
}

@techreport{ref100,
  author      = {Lee, J. and Mukerji, T.},
  title       = {The {Stanford VI-E} reservoir: A synthetic data set for joint seismic--{EM} time-lapse monitoring algorithms},
  institution = {Stanford Center for Reservoir Forecasting, Stanford University},
  type        = {Annual Report},
  number      = {25},
  year        = {2012}
}

@article{Ahmed2026Transformer,
  author = {Ahmed, Nisar and Oikonomou, Dimitrios and Cardozo, Nestor and Weibull, Wiktor and Grana, Dario and Alaei, Behzad and Larsen, Eirik},
  title = {Enhanced Deep Learning Using a Convolutionally Embedded Transformer Architecture for Reservoir Elastic Property Prediction: Comparison With Physics-Based Seismic Inversion},
  journal = {Geophysical Prospecting},
  year = {2026},
  volume = {74},
  number = {5},
  pages = {e70211},
  doi = {10.1111/1365-2478.70211}
}

@article{Ahmed2026Wavelet,
  author = {Ahmed, Ahmed M. and Shragge, Jeffrey and Tsvankin, Ilya},
  title = {Seismic Reservoir Monitoring Using Wavelet Transforms and Machine Learning: A Double-Compression Approach},
  journal = {Applied Sciences},
  year = {2026},
  volume = {16},
  number = {11},
  pages = {5352},
  doi = {10.3390/app16115352}
}

\end{document}